\pdfoutput=1

\documentclass[11pt]{article}
\usepackage[dvipsnames]{xcolor}
\usepackage{ACL2023}

\usepackage{times}
\usepackage{latexsym}

\usepackage[T1]{fontenc}

\usepackage[utf8]{inputenc}

\usepackage{microtype}

\usepackage{inconsolata}
\usepackage{url}
\usepackage{bbm}
\usepackage{tabularx}
\usepackage{graphicx}
\usepackage{booktabs}
\usepackage{algorithm}
\usepackage{algpseudocode}
\usepackage{multirow}
\usepackage{enumitem}
\usepackage{colortbl}
\usepackage{wrapfig,lipsum,booktabs}
\usepackage{bbm}
\usepackage{subfigure}
\usepackage{makecell}
\usepackage{amsmath}
\usepackage{hyperref}
\usepackage{cleveref}
\newcommand{\method}[0]{\textsc{ReConcile}}
\newcommand{\icon}[2][0.4cm]{\begin{minipage}{#1}\includegraphics[width=#1]{#2}\end{minipage}}
\newcommand{\bfull}[0]{\icon{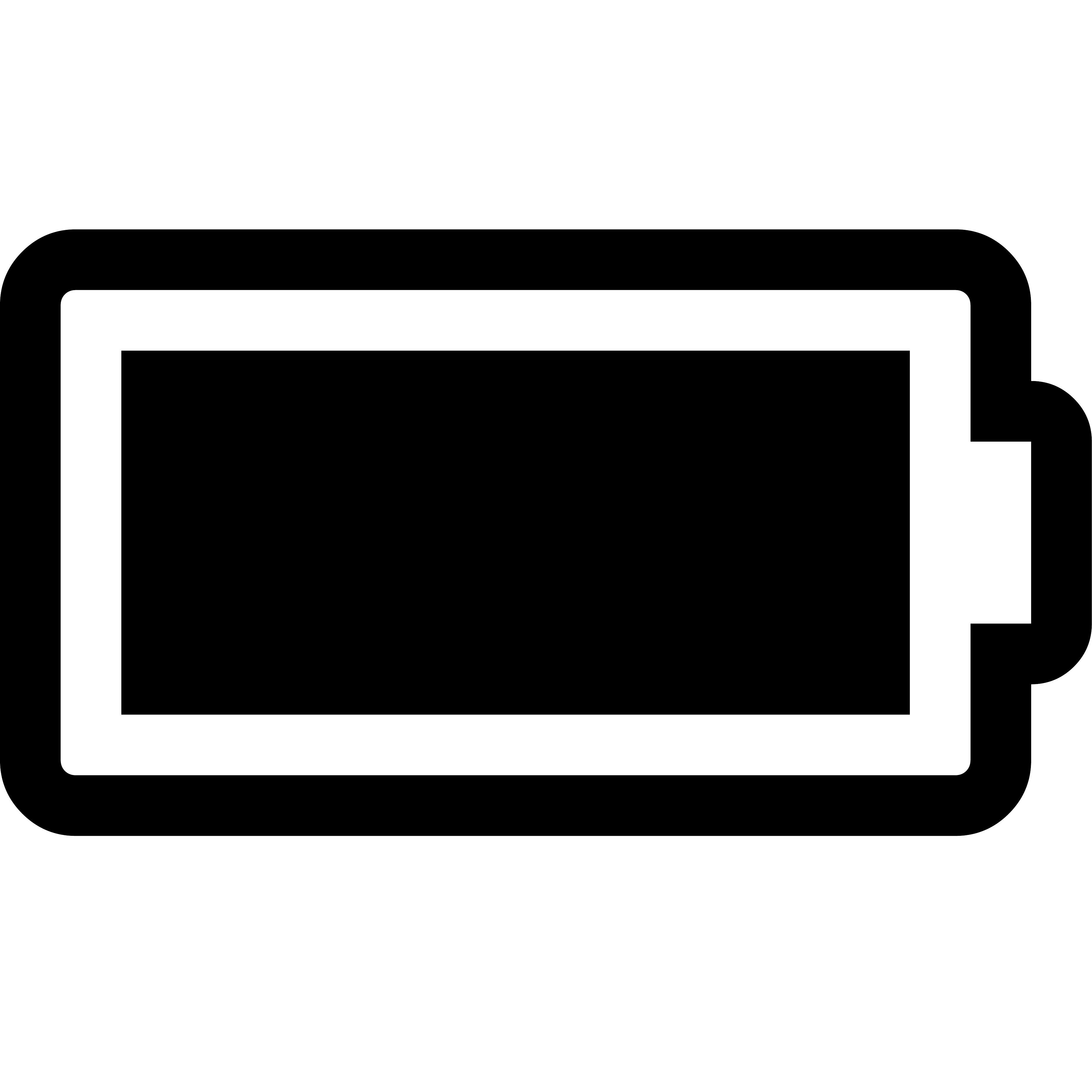}}
\newcommand{\bempty}[0]{\icon{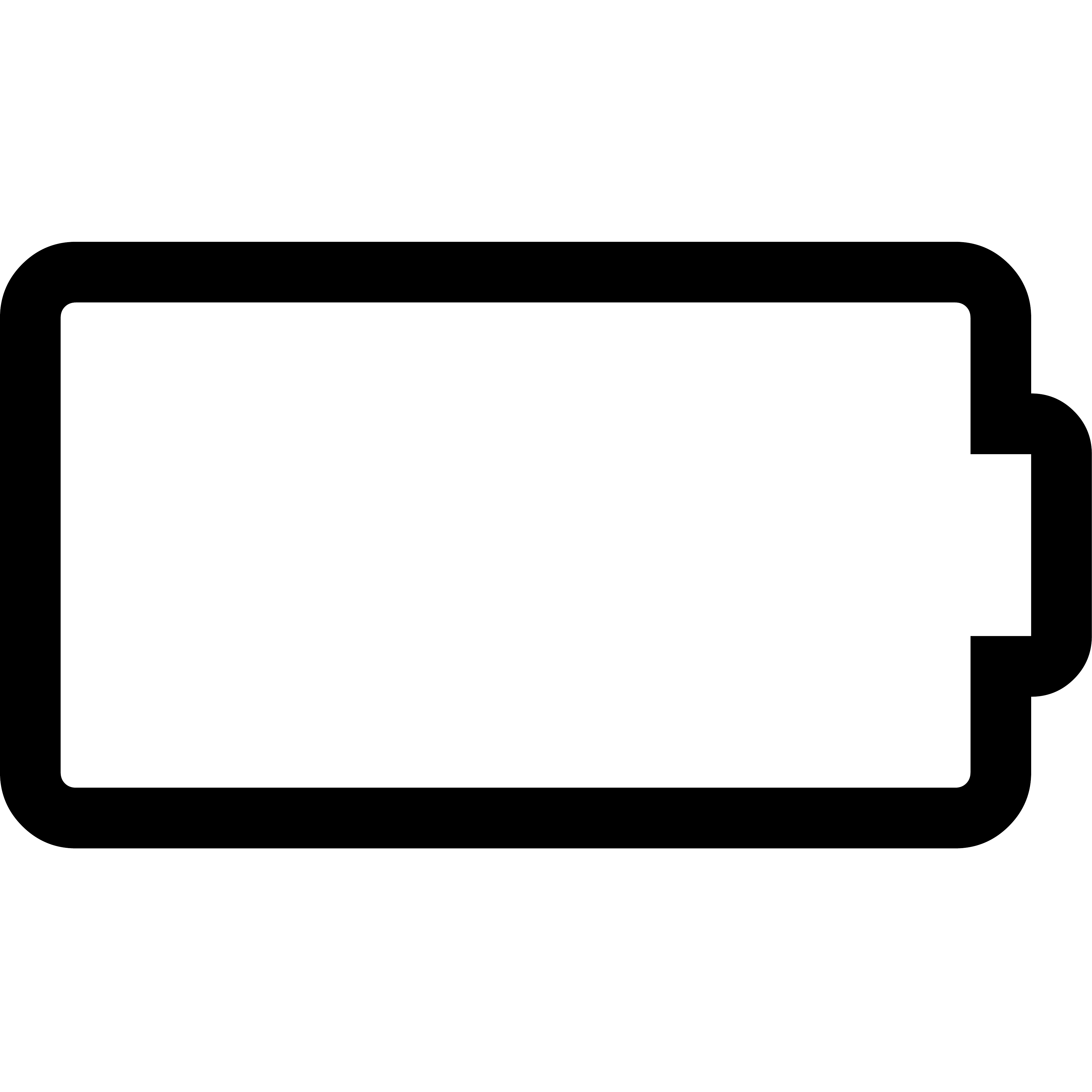}}
\newcommand{\gptfour}[0]{\icon{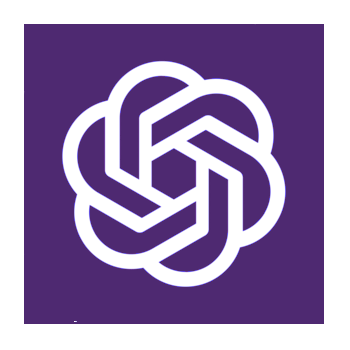}}
\newcommand{\chatgpt}[0]{\icon{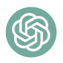}}
\newcommand{\bard}[0]{\icon{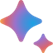}}
\newcommand{\claude}[0]{\icon{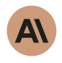}}
\newcommand{\llama}[0]{\icon{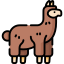}}
\newcommand{\deepseek}[0]{\icon{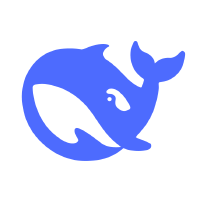}}

\newcommand{\std}[1]{{\tiny$\pm${#1}}}
\algnewcommand\algorithmicforeach{\textbf{for each}}
\algdef{S}[FOR]{ForEach}[1]{\algorithmicforeach\ #1\ \algorithmicdo}
\DeclareMathOperator*{\argmax}{arg\,max}

\title{\method{}: Round-Table Conference Improves Reasoning \\ via Consensus among Diverse LLMs}

\author{Justin Chih-Yao Chen \;\;\;\;\; Swarnadeep Saha \;\;\;\;\; Mohit Bansal \\
        UNC Chapel Hill\\ \texttt{\{cychen,swarna,mbansal\}@cs.unc.edu}}

\begin{document}
\maketitle
\begin{abstract}
Large Language Models (LLMs) still struggle with natural language reasoning tasks. Motivated by the \emph{society of minds}~\citep{minsky1988society}, we propose \method{}, a multi-model multi-agent framework designed as a round table conference among diverse LLM agents. \method{} enhances collaborative reasoning between LLM agents via multiple rounds of discussion, learning to convince other agents to improve their answers, and employing a confidence-weighted voting mechanism that leads to a better consensus. In each round, \method{} initiates discussion between agents via a `discussion prompt' that consists of (a) grouped answers and explanations generated by each agent in the previous round, (b) their confidence scores, and (c) demonstrations of answer-rectifying human explanations, used for convincing other agents.
Experiments on seven benchmarks demonstrate that \method{} significantly improves LLMs' reasoning  -- both individually and as a team -- surpassing prior single-agent and multi-agent baselines by up to 11.4\% and even outperforming GPT-4 on three datasets. \method{} also flexibly incorporates different combinations of agents, including API-based, open-source, and domain-specific models, leading to an 8\% improvement on MATH. Finally, we analyze the individual components of \method{}, demonstrating that the diversity originating from different models is critical to its superior performance.\footnote{Code: \url{https://github.com/dinobby/ReConcile}}
\end{abstract}

\section{Introduction}

\begin{figure*}
    \centering
    \includegraphics[width=0.9\textwidth]{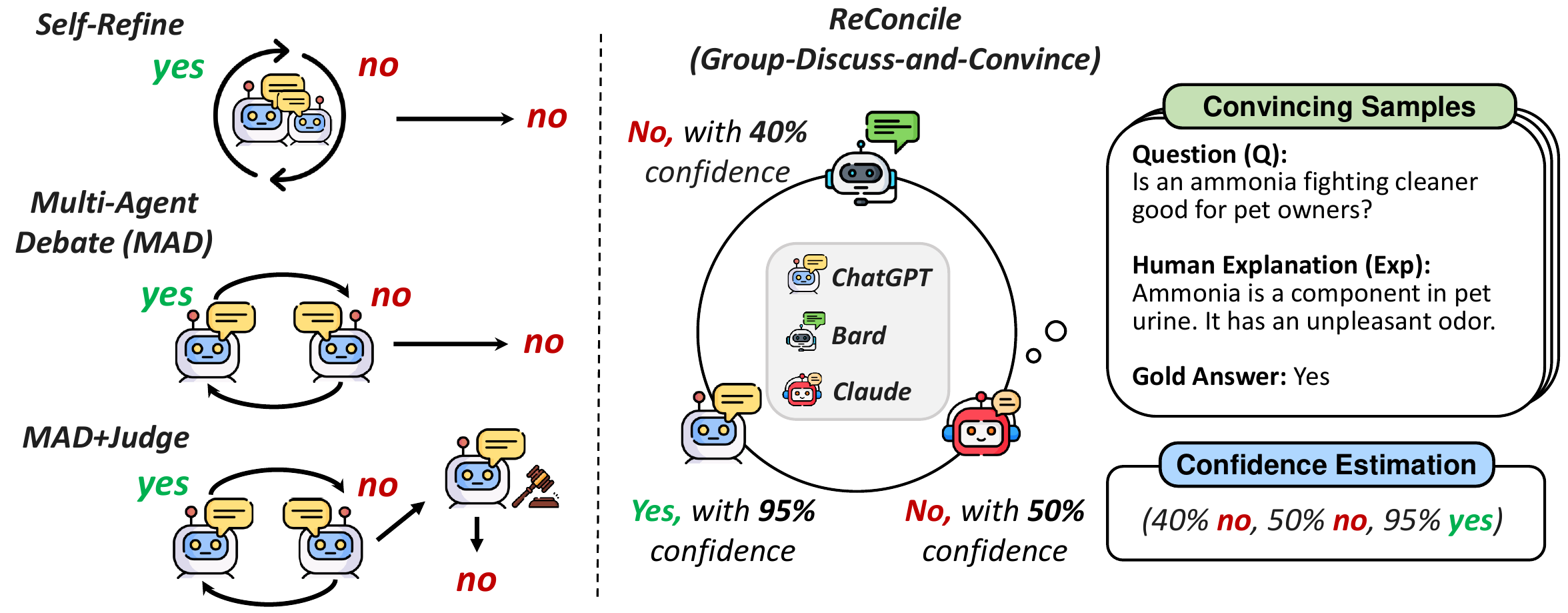}
    \caption{An illustration of the main differences between \method{} and prior works. While most current self-refine and debating techniques rely on multiple instances of a single model (e.g., ChatGPT), our method incorporates models from different families (e.g., ChatGPT, Bard, and Claude2). Our approach also emphasizes critical elements of effective discussion, including convincing another agent to improve their answers and incorporating the estimated confidence of all agents. For illustrative simplicity, we depict only one agent contemplating how to convince the other two agents.
    }
    \label{fig:intro}
\end{figure*}

A large body of recent work has focused on improving the reasoning capabilities of Large Language Models (LLMs) by imitating various human cognitive processes~\citep{wang2023metacognitive,park2023generative,sumers2023cognitive,ye2023cognitive}. These include phenomena like reflecting on and critiquing one's own predictions, being receptive to feedback, and learning from feedback. Of note, self-reflection is an introspective process that allows the model to improve its outputs by generating feedback from the model itself~\citep{madaan2023selfrefine, shinn2023reflexion}. However, self-reflection suffers from Degeneration-of-Thought -- when the model is overly confident in its answer, it is unable to generate novel thoughts even after multiple rounds of feedback~\citep{liang2023encouraging}.

To promote more diverse thoughts, past work has drawn inspiration from the concept of \emph{society of minds} in multi-agent systems~\citep{minsky1988society, zhuge2023mindstorms}. It highlights the importance of communication and collaboration between multiple agents for complex decision-making tasks. 
While such collaborative frameworks like multi-agent debate~\citep{liang2023encouraging, du2023improving} increase the reasoning diversity through the process of a debate, multiple agents have typically been limited to different instances of the same underlying model like ChatGPT~\citep{openai2023gpt3.5}.\footnote{In this work, we refer to multi-agent as multiple instances of the same underlying model (e.g., ChatGPT), whereas multi-model model-agent refers to different models (e.g., ChatGPT, Bard and Claude2) as agents.

}
This results in an inherent model bias, a restricted knowledge scope, and a lack of external feedback from other models due to identical pre-training data and model architectures across all agents. 
In general, when multiple agents propose solutions to a problem, the success of such a multi-agent system is fundamentally reliant on (a) the diversity of the solutions, (b) the ability to estimate each agent's confidence, and (c) accordingly, convince other agents (with explanations) to reach a better consensus. This puts forward the question: if multiple diverse LLMs collaboratively solve a task, are they capable of discussing their solutions with each other to reach a better consensus?

We aim to solve reasoning problems by learning from diverse insights and external feedback, originating from agents that belong to different model families. Collaborative processes such as brainstorming, group meetings, and discussions play a pivotal role in reaching a consensus and arriving at more refined solutions to complex problems~\citep{LI2022790}. Effective discussion also entails the selection of stances, voting, convincing, exchange of information, and a diversity of opinions. Thus, we propose \method{}, a framework of round-table conference for obtaining better consensus among diverse LLM agents. \method{} consists of multiple discussion rounds between diverse LLM agents who try to \emph{convince}\footnote{When we say that an agent tries to convince another agent, we mean that it learns (based on corrective explanations) to defend or argue for its stance while still being receptive to the other agent's argument.} each other to either \emph{rectify} their answers or become more \emph{confident} of their initial correct answers (see Fig.~\ref{fig:intro} for a broad overview). 

\begin{table*}[t]
    \centering
    \small
    \begin{tabular}{lcccccc}
        \toprule
        ~ & Refine & Ensemble & Multi-Agent & Multi-Model & Convincingness & Confidence  \\ 
        \midrule
        Self-Refine (SR) & \bfull & \bempty & \bempty & \bempty & \bempty & \bempty  \\ 
        Self-Consistency (SC) & \bempty & \bfull  & \bempty & \bempty & \bempty & \bempty  \\ 
        SR + SC & \bfull & \bfull  & \bempty & \bempty & \bempty & \bempty  \\ 
        Debate & \bfull & \bfull & \bfull & \hspace{5pt}\bfull {*} & \bempty & \bempty  \\ 
        Judge & \bfull & \bfull & \bfull & \bempty & \bempty & \bempty  \\ 
        \midrule
        \method{} (Ours) & \bfull & \bfull & \bfull & \bfull & \bfull & \bfull  \\ 
        \bottomrule
    \end{tabular}
    \caption{Summary of the main differences between prior work, including Self-Refine (SR, ~\citet{madaan2023selfrefine}); Self-Consistency (SC, ~\citet{wang2022self}); Debate~\citep{du2023improving} and Judge~\citep{liang2023encouraging}. \bfull~means supported and \bempty~means not supported. \method{} supports multi-model multi-agent discussion with confidence estimation and convincingness. * = \citet{du2023improving} primarily experiment with multiple instances of ChatGPT as different agents and conduct an initial investigation with 20 samples using ChatGPT and Bard as the two agents.}
    \label{tab:related_work}
\end{table*}

Given a reasoning problem, \method{} begins with each agent first generating an answer, its uncertainty, and a corresponding explanation (as a Chain-of-Thought~\citep{wei2022chain}) for the answer. Then all agents enter a multi-round discussion phase. Each discussion round consists of all agents generating a revised explanation and answer based on all other agents' explanations and answers from the previous round. In particular, \method{} initiates a discussion by designing a \emph{discussion prompt} for each agent, that lets it condition on (1) grouped answers from all agents, (2) corresponding explanations generated in the previous round, and (3) demonstrations of answer-rectifying human explanations for convincing other agents. 
We leverage them in an in-context learning framework to teach models to generate their own convincing explanations (see Fig.~\ref{fig:convincing_samples_choosing}). Even in cases where an agent initially offers an incorrect answer and explanation, it can consider another agent's convincing explanation and amend its response accordingly. In each discussion round, we estimate an agent's uncertainty via a confidence-estimation prompt~\citep{tian2023just, xiong2023examining}. Once all agents converge to the same answer (i.e., a consensus has been reached), we employ these confidences to compute a weighted vote as the team answer. 

We primarily develop \method{} with three state-of-the-art LLMs: ChatGPT~\citep{openai2023gpt3.5}, Bard~\citep{palm2}, and Claude2~\citep{claude2}. We also demonstrate the flexibility of \method{} with variants that employ a much stronger GPT-4~\cite{openai2023gpt4}, an open-source LLaMA-2-70B~\cite{touvron2023llama}, or a domain-specific DeepSeekMATH~\cite{deepseek-math} model as an agent. Across seven benchmarks spanning commonsense reasoning, mathematical reasoning, logical reasoning, and Natural Language Inference (NLI), \method{} outperforms prior single-agent (e.g., Self-Refine~\citep{madaan2023selfrefine} and Self-consistency~\citep{wang2022self}) and multi-agent baselines (Debate~\citep{du2023improving} and Judge~\citep{liang2023encouraging}) that are built on top of the same underlying models. For example, \method{}, (1) on a date understanding task, outperforms the leading multi-agent debate baseline by 11.4\%, (2) on StrategyQA, also outperforms GPT-4 by 3.4\%, and (3) on MATH, outperforms both GPT-4 and a specialized DeepSeekMath model by 8\%. Moreover, detailed analyses of the individual components of \method{} demonstrate that leveraging diverse LLM agents leads to maximum gains, and we further validate their higher response diversity via a BERTScore-based diversity metric~\citep{zhang2019bertscore}. Finally, we show that \method{} not only leads to better team performance but also enables each agent to improve individually via the discussion process.

In summary, our primary contributions are:
\begin{itemize}[nosep, wide=0pt, leftmargin=*, after=\strut]
  \item We propose \method{}, a reasoning framework involving diverse Large Language Models in a Round Table Conference.
  \item We conduct extensive experiments on seven benchmarks to show that \method{} outperforms strong baselines (including GPT-4 on some benchmarks) and also generalizes to different combinations of agents.
  \item We study the role of diversity, confidence estimation, and an agent's ability to convince others (by learning from corrective explanations) in multi-agent discussion systems. 
\end{itemize}

\begin{figure*}[t]
\centering
\includegraphics[width=\textwidth]{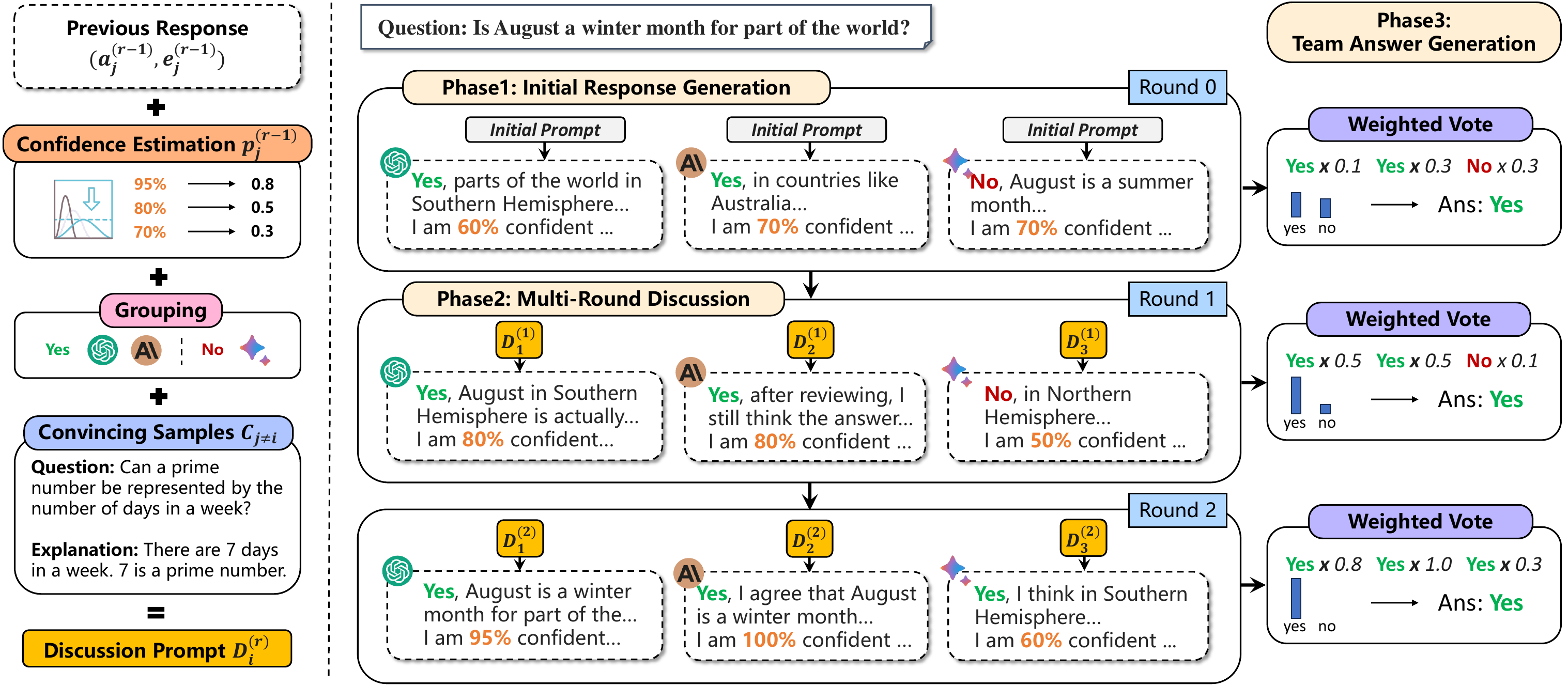}
\caption{Overview of \method{} with ChatGPT, Bard, and Claude2, consisting of three phases: (1) Initial Response Generation:  Each agent generates an initial answer and explanation. (2) Multi-Round Discussion: Each model is presented with a discussion prompt (as illustrated on the left) and subsequently generates an updated answer and explanation. 
(3) Team answer generation: The team answer is determined by a weighted vote at the end of each round. The left part of the figure shows the discussion prompt for an agent, consisting of (a) grouped answers and explanations of all agents from the previous round, (b) estimated confidence, and (c) demonstrations of convincing samples. 
}
\vspace{-5pt}
\label{fig:overview}
\end{figure*}

\section{Related Work}

\noindent \textbf{Reasoning with LLMs.}
Progress in LLMs has led to the development of advanced prompting and fine-tuning techniques for solving reasoning problems. Representative methods include Chain-of-Thought (CoT)~\citep{kojima2022large, wei2022chain, wang-etal-2023-plan} and Tree-of-Thought prompting~\citep{yao2023tree}, self-consistency~\citep{wang2022self}, meta-reasoning over multiple paths~\citep{yoran2023answering}, use of scratchpads~\citep{nye2021show}, training verifiers~\citep{cobbe2021training}, self-collaboration~\citep{wang2023unleashing,schick2022peer,li2023camel,feng2024dont}, self-reflection~\citep{shinn2023reflexion, madaan2023selfrefine, wang2023metacognitive,yao2023react}, improved math reasoning~\citep{yue2023mammoth,luo2023wizardmath} and fine-tuning via bootstrapping models~\citep{zelikman2022star, lewkowycz2022solving, li-etal-2023-making}. Eliciting reasoning from a single agent, while promising, is fundamentally limited by a lack of diverse insights. 

\noindent \textbf{Reasoning in Multi-Agent Systems.}
A recent line of work has explored student-teacher frameworks with the goal of distilling reasoning capabilities from a stronger teacher to a weaker student~\citep{magister2022teaching, fu2023specializing, ho2022large, saha2023can, mukherjee2023orca}. As opposed to a teacher teaching weaker agents, we seek to develop a multi-agent system where different LLM agents have their unique strengths and try to collaboratively improve performance by reaching a better consensus. Notable prior works include multi-agent debating frameworks~\citep{du2023improving, liang2023encouraging, chan2023chateval, xiong2023examining, khan2024debating} but such efforts are still largely limited to multiple instances of the same underlying language model. We argue that relying on a single model limits the potential of complementary benefits from different model families and the advantage of ensemble learning. 
Moreover, estimating the confidence of each agent and being able to defend or improve one's opinions become more prominent components in such multi-model multi-agent systems because of the individual differences. Overall, Table~\ref{tab:related_work} summarizes \method{}'s key differences compared to prior single-agent and multi-agent reasoning methods.

\looseness-1
\noindent \textbf{Ensembling Large Pretrained Models.}
Large pre-trained models, by virtue of being trained on different data and with architectural variations, exhibit distinct capabilities. This has led to the development of ensembles~\citep{sagi2018ensemble} in multimodal learning~\citep{zeng2022socratic, li2022composing}. Mixture of Experts, a popular ensemble learning technique, trains multiple smaller specialized models to improve robustness and overall accuracy~\citep{jacobs1991adaptive, shazeer2017outrageously, du2022glam}. Specific to language models, Self-Consistency~\citep{wang2022self} generates diverse reasoning paths using CoT and chooses the most consistent answer as the final output. 
\citet{jiang-etal-2023-llm} propose LLM-Blender, a method to rank and fuse generations from different models. Different from these, we study communication via explanations between distinct LLM agents and their ability to discuss and convince each other in order to improve collective reasoning.
\section{Problem Setup}

\looseness-1
We assume that we are given a test problem $Q$ and there are $n$ agents $\mathcal{A} = \{A_i\}_{i=1}^{n}$ participating in a round table discussion. Each agent is a distinct LLM, potentially trained with different pre-training data and model architectures. All agents are capable of generating an answer and a corresponding Chain-of-Thought explanation~\citep{wei2022chain} for the test problem. For each agent $A_i$, we utilize a small number of $k$ demonstrations of convincing samples $C_i = \{c^{(i)}_j\}_{j=1}^k$. Each convincing sample $c^{(i)}_j = (q^{(i)}_j, a^{(i)}_j, e^{(i)}_j)$ for an agent $A_i$ is an instance of a question $q^{(i)}_j$, gold answer $a^{(i)}_j$, and a human explanation $e^{(i)}_j$ that helps rectify an agent's initial incorrect answer (see more details in Sec~\ref{sec:phase2}). The objective of \method{} is to improve the team performance on a given task by holding multiple rounds of discussion between the agents, quantifying the uncertainty associated with each agent, and convincing other agents to reach a better consensus. Note that convincing samples serve as an additional performance enhancer; even when the dataset lacks human explanations, our method can still yield performance gains independent of this (more details below).

\section{\method{}: A Collaborative Discussion Framework}
\label{sec:method}

\method{} operates in three phases: initial response generation, multi-round discussion, and team answer generation. The overview of our method is demonstrated in Fig.~\ref{fig:overview} and Algorithm~\ref{alg:RTC}.

\paragraph{Phase 1: Initial Response Generation.}
\method{} operates with each agent $A_i$ initially generating an answer $a_i^{(0)}$, an explanation $e_i^{(0)}$, and an associated confidence $p_i^{(0)}\in[0, 1]$ for the generated answer. Each agent conditions on a zero-shot prompt that instructs it to reason about the problem `step-by-step'. See `Phase 1' in Fig.~\ref{fig:overview} and the prompt is shown in Fig.~\ref{fig:prompt} in Appendix~\ref{appendix:prompt}. 

\paragraph{Phase 2: Multi-round Discussion.}
\label{sec:phase2}
\method{} then enters a discussion phase, consisting of $R$ rounds (see `Phase 2' in Fig.~\ref{fig:overview}). In discussion round $r$, for each agent $A_i$, \method{} develops a discussion prompt $\mathcal{D}^{(r)}_i$ (as shown in Fig.~\ref{fig:prompt}), consisting of the following three components.

\noindent \textbf{(a) Grouped responses of all agents from the previous round.} $\mathcal{D}^{(r)}_i$ consists of the answers $\{a_j^{(r-1)}\}_{j=1}^n$ and explanations $\{e_j^{(r-1)}\}_{j=1}^n$ of all agents from round $(r-1)$. To foster better discussions, \method{} summarizes this information by grouping the answers into distinct categories and appends all plausible explanations for each answer, as shown in our discussion prompt (Appendix Fig.~\ref{fig:prompt}) and on the left side of Fig.~\ref{fig:overview}.

\noindent \textbf{(b) Confidence associated with the answers.} All agents are not equally confident in their answers. Hence, an effective discussion should also consider each agent's uncertainty. For all black-box models, we estimate its confidence $p_i^{(r)}$ in round $r$ by directly prompting the agent to verbally quantify its uncertainty, which in past work has been shown to be effective~\citep{xiong2023can}. See Appendix Fig.~\ref{fig:prompt} for the usage of confidence in discussion.

\begin{figure}[t]
\centering
        \includegraphics[width=\columnwidth]{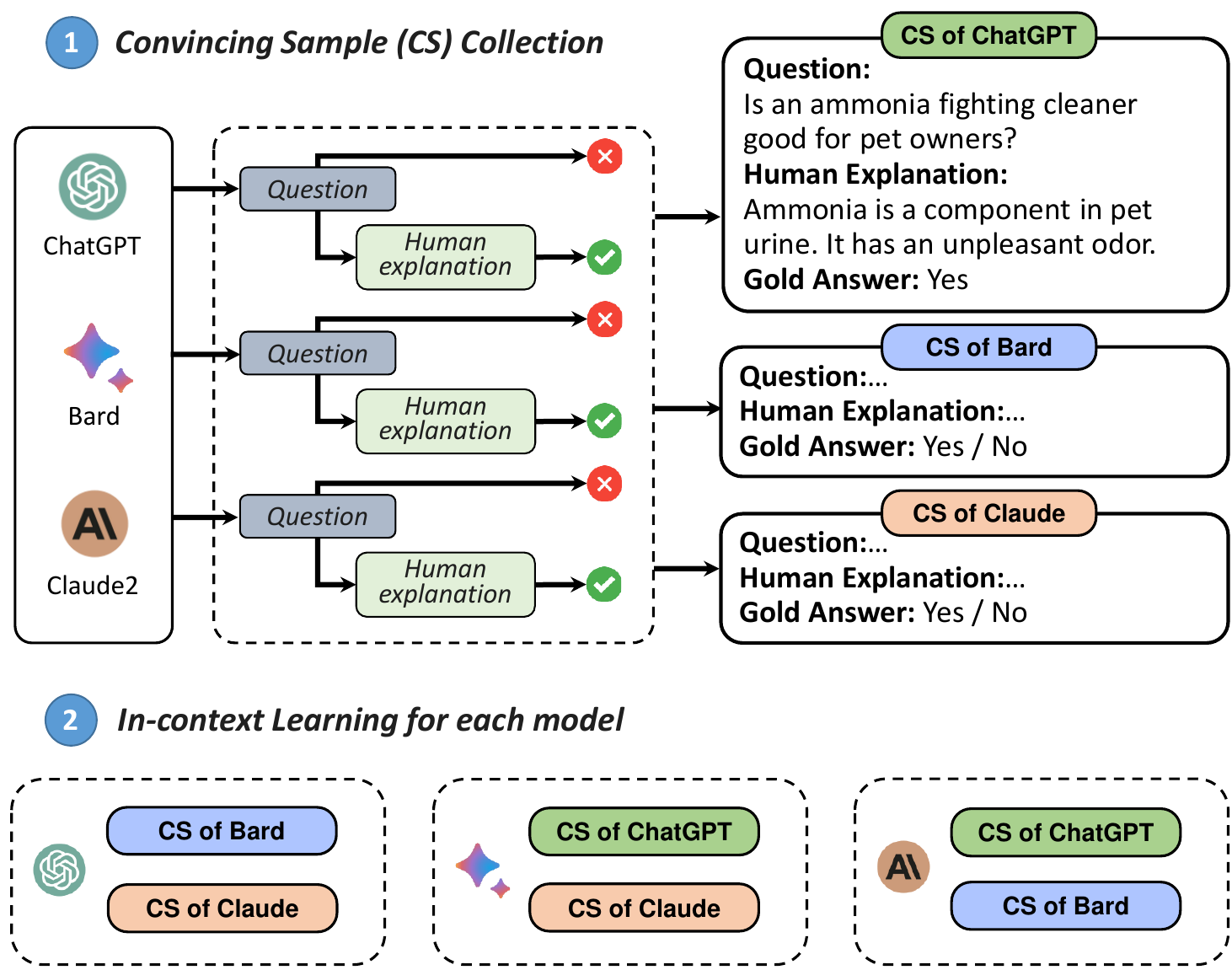}
        \caption{\label{fig:convincing_samples_choosing}Method for choosing convincing samples for each agent. 
        A convincing sample for ChatGPT consists of a question, a gold answer, and a `corrective' human explanation that can rectify its initial incorrect answer. Then Bard and Claude2 use it in-context during discussion to convince ChatGPT.
        } 
\end{figure}
\noindent \textbf{(c) Convincing samples from all other agents.} Finally, the prompt contains convincing samples $C_j$ for all other agents $A_{j \neq i}$.\footnote{We did not include an agent's own convincing samples in the prompt because an agent is expected to specifically convince \emph{other} agents. We also verify this empirically -- additionally including self-convincing samples in the prompt leads to comparable performance.} When an agent tries to reassess its reasoning in light of the reasoning provided by other agents, we hypothesize that it should benefit from conditioning on demonstrations that can convince other agents. In order to obtain such convincing samples for an agent $A_j$, we select a small number of samples (4 in our experiments) for which the agent's initial answer is wrong but conditioning on the corresponding human explanation, rectifies the answer (see Fig.~\ref{fig:convincing_samples_choosing}). For datasets that \emph{do not} come with human explanations (e.g., the date understanding task in our experiments), we develop \method{} without using any convincing sample in the discussion prompt and still obtain large improvements (see \S\ref{sec:ablation} for details).

We now define the discussion prompt~$\mathcal{D}_i^{(r)} = \{a_j^{(r-1)}, e_j^{(r-1)}, p_j^{(r-1)}, C_{j \neq i}\}_{j=1}^n$ for each agent $A_i$ in round $r$, based on the above three components. The agent conditions on it to generate an updated answer $a_i^{(r)}$, explanation $e_i^{(r)}$, and confidence $p_i^{(r)}$, to be used in the next round. 
Demonstrations of convincing explanations enable the agent to generate explanations that are more likely to convince other agents to reach a better consensus. 

\paragraph{Phase 3: Team Answer Generation.} 
\method{} continues the discussion for a maximum of $R$ rounds or terminates it as soon as a consensus is reached (i.e., all agents agree on the same answer). At the end of any round $r$, \method{} generates the team answer $\hat{a}^{(r)}$ for that round using a weighted voting scheme (see the right side of Fig.~\ref{fig:overview}). In particular, we recalibrate each agent's confidence using a function $f(\cdot)$ and then use these as weights to compute the team answer, as follows:
\begin{equation}
        \hat{a}^{(r)} = \argmax_a\sum_{i}{f(p_i^{(r)})}\mathbbm{1}(\hat{a}^{(r)}_i=a) \nonumber
\end{equation}
where $a$ is a distinct answer generated by any of the agents, $p_i^{(r)}$ is the original confidence of agent $A_i$ in round $r$ and $f(p_i^{(r)})$ is the corresponding recalibrated confidence. While an unweighted majority vote and uncalibrated confidence-weighted vote also work well in practice, we use the calibrated weighted vote because it not only obtains slightly better results but the same recalibration strategy also works out-of-the-box for all seven tasks that we experiment with (see Appendix~\ref{sec:vote_strategies} for more details of our recalibration function $f(\cdot)$).
\section{Experimental Setup}
\textbf{Agents in \method{}.}
We primarily implement \method{} with ChatGPT, Bard, and Claude2 as the three agents, engaging them in up to three rounds of discussion. Later in \S\ref{sec:main_result}, we also show the generalizability of our \method{} framework with different choices of agents, including API-based (GPT-4), open-source (LLaMA-2-70B), and domain-specific (DeepSeekMath) agents.

\noindent\textbf{Datasets.}
We evaluate \method{} on seven benchmarks, including two commonsense, three math, one logical reasoning, and one NLI task. These are: (1) StrategyQA~\citep{geva2021did}, (2) CommonsenseQA (CSQA;~\citep{aggarwal2021explanations, talmor-etal-2019-commonsenseqa}), (3) GSM8K~\citep{cobbe2021training}, (4) AQuA~\citep{ling2017program}, (5) MATH~\citep{hendrycksmath2021}, (6) Date Understanding~\citep{srivastava2023beyond}, and (7) ANLI~\citep{nie-etal-2020-adversarial}.

\noindent\textbf{Baselines.}
We compare \method{} to prior works in three categories: 
\begin{itemize}[nosep, wide=0pt, leftmargin=*, after=\strut]
    \item \textbf{Vanilla single-agent methods.} In this category, we experiment with (1) zero-shot CoT prompting~\citep{kojima2022large} with one of the interacting LLMs, and (2) eight-shot CoT with Claude2 where the number eight matches the number of convincing samples used in \method{}.
    \item \textbf{Advanced single-agent methods.} Next, we compare with (1) Self-Refine (SR) that iteratively generates feedback and refines the output leveraging the model itself~\citep{madaan2023selfrefine}, (2) Self-Consistency (SC) that samples multiple reasoning paths and generates the most consistent answer~\citep{wang2022self}, and (3) their combination, SR+SC, that first conducts multiple iterations of refinement, followed by a majority vote. Note that in \method{}, the number of LLM calls per instance can vary between 3, 6, and 9 based on the number of discussion rounds. Hence, for a fair comparison, we implement SC with the same average number of LLM calls as in \method{}. Later in Appendix~\ref{sec:nine_way_SC}, we show that \method{} even outperforms 9-way SC (that equates to the worst-case LLM calls in \method{}). 
    \item \textbf{Multi-agent methods with a single backbone model.} Our final baselines are two multi-agent debating methods: a multi-agent debate between multiple ChatGPT instances~\citep{du2023improving} and a debate with judge method~\citep{liang2023encouraging}. These methods use multiple instances of the same underlying model (ChatGPT) as different agents.
\end{itemize}

\noindent\textbf{Implementation Details}.
Owing to the cost associated with API-based models and the limit imposed on the number of API calls, we follow many prior works~\citep{du2023improving,bian2023chatgpt,besta2023graph,yao2023tree} to experiment with a subset of 100 samples (from the validation set for StrategyQA and the test set for all other datasets). Later in Appendix~\ref{sec:full_test}, we also experiment on the full test sets of StrategyQA and Date understanding and find similar trends.
We report accuracy and its standard deviation. For each experiment, we conduct at least three runs on the same test samples with the same prompts, primarily accounting for the variance caused by the decoding strategy. Other implementation details can be found in Appendix~\ref{appendix:implementation}.

\begin{table*}[t]
\small
    \centering
    \resizebox{\textwidth}{!}{%
    \begin{tabular}{lllccccccc}
        \toprule
         Method Category &  Method & Agent & StrategyQA & CSQA & GSM8K & AQuA & Date\\ \midrule
        \multirow{5}{2cm}{Vanilla Single-agent} 
         & Zero-shot CoT & \gptfour~GPT-4 & \color{gray} \textit{75.6}\std{4.7} & \color{gray} \textit{73.3}\std{0.4} & \color{gray} \textit{90.7}\std{1.7} & \color{gray} \textit{65.7}\std{4.6} & \color{gray} \textit{89.0}\std{2.2} \\
        & Zero-shot CoT & \chatgpt~ChatGPT & 67.3\std{3.6} & 66.0\std{1.8} & 73.7\std{3.1} & 44.7\std{0.5} & 67.7\std{1.2} \\ 
        & Zero-shot CoT & \bard~Bard & 69.3\std{4.4} & 56.8\std{2.7} & 58.7\std{2.6} & 33.7\std{1.2} & 50.2\std{2.2} \\
        & Zero-shot CoT & \claude~Claude2 & 73.7\std{3.1} & 66.7\std{2.1} & 79.3\std{3.6} & 60.3\std{1.2} & 78.7\std{2.1}   \\
        & Eight-shot CoT & \claude~Claude2 & 74.3\std{0.8} & 68.3\std{1.7} & 84.7\std{0.9} & 64.7\std{1.2} & 78.7\std{1.7}   \\\midrule
        \multirow{3}{2cm}{Advanced Single-agent} & Self-Refine (SR) & \chatgpt~ChatGPT & 66.7\std{2.7} & 68.1\std{1.8} & 74.3\std{2.5} & 45.3\std{2.2}  & 66.3\std{2.1} \\ 
        & Self-Consistency (SC) & \chatgpt~ChatGPT & 73.3\std{0.5} & 73.0\std{0.8} & 82.7\std{0.5} & 60.3\std{1.2} & 69.3\std{0.4} \\
        & SR + SC & \chatgpt~ChatGPT & 72.2\std{1.9} & 71.9\std{2.1} & 81.3\std{1.7} & 58.3\std{3.7} & 68.7\std{1.2}  \\ \midrule
        \multirow{4}{2cm}{Single-model Multi-agent} 
        & Debate & \chatgpt $\times 3 $ & 66.7\std{3.1} & 62.7\std{1.2} & 83.0\std{2.2} & 65.3\std{3.1}  & 68.0\std{1.6}  \\ 
        & Debate & \bard $\times 3 $ & 65.3\std{2.5} & 66.3\std{2.1} & 56.3\std{1.2} & 29.3\std{4.2} & 46.0\std{2.2}\\
        & Debate & \claude $\times 3 $ & 71.3\std{2.2} & 68.3\std{1.7} & 70.7\std{4.8} & 62.7\std{2.6} & 75.3\std{3.3}\\

        & Debate+Judge & \chatgpt $\times 3 $ & 69.7\std{2.1} & 63.7\std{2.5} & 74.3\std{2.9} & 57.3\std{2.1} & 67.7\std{0.5} \\ \midrule
        \multirow{1}*{Multi-model Multi-agent} & \method{} & \chatgpt, \bard, \claude & \textbf{79.0}\std{1.6} & \textbf{74.7}\std{0.4} & \textbf{85.3}\std{2.2} & \textbf{66.0}\std{0.8} & \textbf{86.7}\std{1.2} \\ 
        \bottomrule
    \end{tabular}
    }
    \caption{
    Comparison of \method{} (using ChatGPT, Bard, Claude2) with vanilla and advanced single-agent methods and multi-agent debating frameworks. Across all reasoning benchmarks, \method{} outperforms all prior single-agent and multi-agent methods. On commonsense tasks (StrategyQA and CSQA), \method{} also outperforms GPT-4. All results are on a random subset of 100 samples.
    The agents are \gptfour~GPT-4, \chatgpt~ChatGPT, \bard~Bard, and \claude~Claude2.
    }
    \vspace{-10pt}
    \label{table:main_table}
\end{table*}

\begin{table}[t]
\centering
    \resizebox{\columnwidth}{!}{%
\small
\begin{tabular}{lcc}
\toprule
                    Method & \multicolumn{2}{c}{Accuracy} \\\midrule
Best Single-agent (zero-shot)   & 75.6 (\gptfour) & 73.7 (\claude)\\
Best Multi-agent (Debate)    & 83.7 (\gptfour $\times 3$) & 71.3 (\chatgpt $\times 3$) \\
\textsc{\method{}}           & \textbf{87.7} (\gptfour,\bard,\claude) & \textbf{78.0} (\chatgpt,\claude,\llama) \\
\bottomrule
\end{tabular}}
\caption{Comparison of the best single-agent, best multi-agent, and \method{} on StrategyQA for a given combination of three agents. \method{} flexibly incorporates agents with varying strengths, such as a stronger model like \gptfour~GPT-4, or an open-source model like \llama~LLaMA2-70B.}
\vspace{-10pt}
\label{tab:flexibility_strength}
\end{table}

\section{Results}
\subsection{Main Results}
\label{sec:main_result}
\paragraph{\method{} outperforms single-agent and multi-agent baselines.}
We first evaluate the overall reasoning capabilities of \method{} in Table~\ref{table:main_table} with ChatGPT, Bard, and Claude2 as the three agents. For fair comparisons, all iterative methods go through 3 rounds of iteration and all single-model multi-agent baselines are implemented with three agents with a sufficiently high temperature of 1.0 for maximizing diversity. 
Across all five datasets, \method{} outperforms all single-agent and multi-agent baselines that are built on top of the same models (see last row). Notably, without using GPT-4 as an agent, our method outperforms GPT-4 on commonsense tasks like StrategyQA and CSQA and obtains comparable performance to GPT-4 on most other tasks. 
GPT-4's especially strong results on GSM8K could be attributed in part to the inclusion of some of GSM8K's training samples in GPT-4's pre-training data~\citep{openai2023gpt4}.
While multi-agent debate with ChatGPT~\citep{du2023improving} improves results on math benchmarks, debate with multiple Bard or Claude2 instances is not effective, possibly because the responses (generated from the same model) are not sufficiently diverse. When they team up with ChatGPT in a multi-round discussion, \method{} outperforms debate frameworks. It obtains maximum gains of 11.4\% (75.3\% $\rightarrow$ 86.7\%) on date understanding and 7.7\% (71.3\% $\rightarrow$ 79.0\%) on StrategyQA when compared to the strongest baseline (multi-agent debate with Claude2). Improvements in the math reasoning tasks are relatively moderate, because of ChatGPT's initial strong performance. 
However, as demonstrated later in Table~\ref{tab:flexibility_domain}, integrating a specialized math reasoning model into \method{} significantly boosts team performance.

\paragraph{\method{} generalizes to agents of varying strengths.}
Next, we vary the agents in \method{} to study its generalization as a multi-agent framework. In particular, we either include (a) a stronger GPT-4 model, or (b) an open-source LLaMA-2-70B-chat model in the discussion. As shown in Table~\ref{tab:flexibility_strength}, in both these scenarios, \method{} outperforms the best single-agent and multi-agent baselines, notably even outperforming the zero-shot GPT-4 performance by 12.1\% (75.6\% $\rightarrow$ 87.7\%) on StrategyQA. This highlights the potential of a stronger agent to also obtain useful external feedback from comparatively weaker agents.

\begin{table}[t]
\centering
\small
\begin{tabular}{lc}
\toprule
                   Method & Accuracy \\\midrule
GPT-4 (zero-shot)              & 44.0 (\gptfour) \\
Best Single-agent (zero-shot)  & 50.5 (\deepseek) \\
Best Multi-agent (Debate)   & 48.7 (\gptfour $\times 3$) \\
\textsc{\method{}}  & \textbf{58.3} (\gptfour,\claude,\deepseek) \\
\bottomrule
\end{tabular}
\caption{\method{} generalizes to specialized models like \deepseek~DeepSeekMath and improves on a challenging mathematical reasoning benchmark, MATH.}
\label{tab:flexibility_domain}
\vspace{-10pt}
\end{table}

\begin{table*}[t]
\centering
\small
\begin{tabular}{llccccc}
\toprule
Metric & Method & Accuracy & D (A1, A2) & D (A1, A3) & D (A2, A3) & D (A1, A2, A3) \\ \midrule
\multirow[c]{3}{*}{BERTScore} & \method{} (\chatgpt~Paraphrased) & 72.2  &  0.9364  &   0.9376   &    0.9453   &    0.9398   \\
                        & \method{} (\chatgpt~$\times 3$) & 72.2 & 0.9077 & 0.9181 & 0.9049 & 0.9102 \\ 
                        & \method{} (\chatgpt, \bard, \claude) & \textbf{79.0} & \textbf{0.8891} & \textbf{0.8833} & \textbf{0.8493} & \textbf{0.8739} \\ \bottomrule
\end{tabular}
  \vspace{-5pt}
\caption{Comparison of diversity between (a) paraphrased responses (first row) and (b) responses from multiple instances of the same ChatGPT model (second row). \method{} with a multi-model component also leads to higher accuracy. Responses from different models in \method{} (last row) are most diverse (i.e., less similar).}
  \vspace{-5pt}

\label{table:diversity}
\end{table*}

\paragraph{\method{} generalizes to domain-specific agents.}
So far, we have experimented with \method{} variants that employed general-purpose models like ChatGPT as agents. Our next result in Table~\ref{tab:flexibility_domain} shows that even for tasks that require substantial domain knowledge (e.g., the MATH benchmark~\cite{hendrycksmath2021}), \method{} is flexible enough to utilize and improve upon specialized, domain-specific models. Recently, \citet{deepseek-math} proposed DeepSeekMath, a 7B model pre-trained on a large number of math-related web corpus and improving over GPT-4. Notably, \method{} with GPT-4, Claude2, and DeepSeekMath as agents significantly outperforms zero-shot DeepSeekMath and GPT4-based Debate by 7.8\% and 9.6\% respectively. In summary, \method{} shows consistent improvements across a wide range of agent combinations (involving API-based, open-source, and domain-specific models).

\paragraph{\method{} also improves Natural Language Inference.}
While all our previous results were with reasoning tasks, we also demonstrate \method{}'s effectiveness on ANLI~\citep{nie-etal-2020-adversarial}, a challenging Natural Language Inference benchmark. Table~\ref{tab:flexibility_tasks} shows that \method{} on ANLI outperforms Debate by a significant 9.4\%, pointing to its widespread applicability.

\begin{table}[t]
\centering
\small
\begin{tabular}{lc}
\toprule
                  Method  & Accuracy \\\midrule
Best Single-agent (zero-shot)  & 51.3 (\claude) \\
Best Multi-agent (Debate)    & 48.3 (\chatgpt $\times 3$) \\
\textsc{\method{}}           & \textbf{57.7} (\chatgpt,\bard,\claude) \\
\bottomrule
\end{tabular}
\caption{\method{} improves a challenging NLI benchmark (ANLI), outperforming Debate by 9.4\%.}
\label{tab:flexibility_tasks}
\end{table}

\subsection{Ablations and Analysis of \method{}}
\label{sec:ablation}
\textbf{Each component of \method{} improves reasoning.} In Table~\ref{table:ablation}, we evaluate individual components of \method{} on StrategyQA. In particular, we compare four variants: (1) \textbf{w/o Multiple Models}: We use ChatGPT as the backbone for all three agents, (2) \textbf{w/o Grouping}: We simply concatenate the responses from different agents without grouping their answers, (3) \textbf{w/o Convincingness}: We remove convincing samples from all prompts, and (4) \textbf{w/o Confidence Estimation}: We do not use any confidence estimates during the discussion and compute majority vote as the team answer. We show that each component has a positive impact on \method{} with varying capacities. The effect of different models as agents is particularly significant and we observe a 6.8\% improvement compared to only using ChatGPT as all three agents. This reinforces our hypothesis (and further verified below in `Diversity Analysis') that diverse LLMs have complementary strengths and when put together in a round table discussion, they can learn from diverse external feedback from other agents and refine their responses to reach a better consensus. Notably, convincing samples lead to a 4.5\% improvement in accuracy. In Appendix~\ref{appendix:convincing_ablation}, we study the role of convincing samples to show that (1) they also improve other interaction frameworks, and (2) even in the absence of such examples, \method{} outperforms debate baselines. 

\begin{table}[t]
\small
    \centering
    \begin{tabular}{lcccccc}
        \toprule
        Method & Accuracy  \\ 
        \midrule
        \method{} & \textbf{79.0}\std{1.6} \\
        w/o Multiple Models & 72.2\std{2.1} \\
        w/o Grouping & 76.7\std{2.5} \\
        w/o Convincingness & 74.5\std{1.7} \\
        w/o Conf Estimation & 77.7\std{1.3} \\
        \bottomrule
    \end{tabular}
    \caption{Ablations of \method{} on StrategyQA.}
    \vspace{-10pt}
    \label{table:ablation}
\end{table}

\begin{figure*}[t]
\centering
  \subfigure[]{\includegraphics[width=0.329\textwidth]{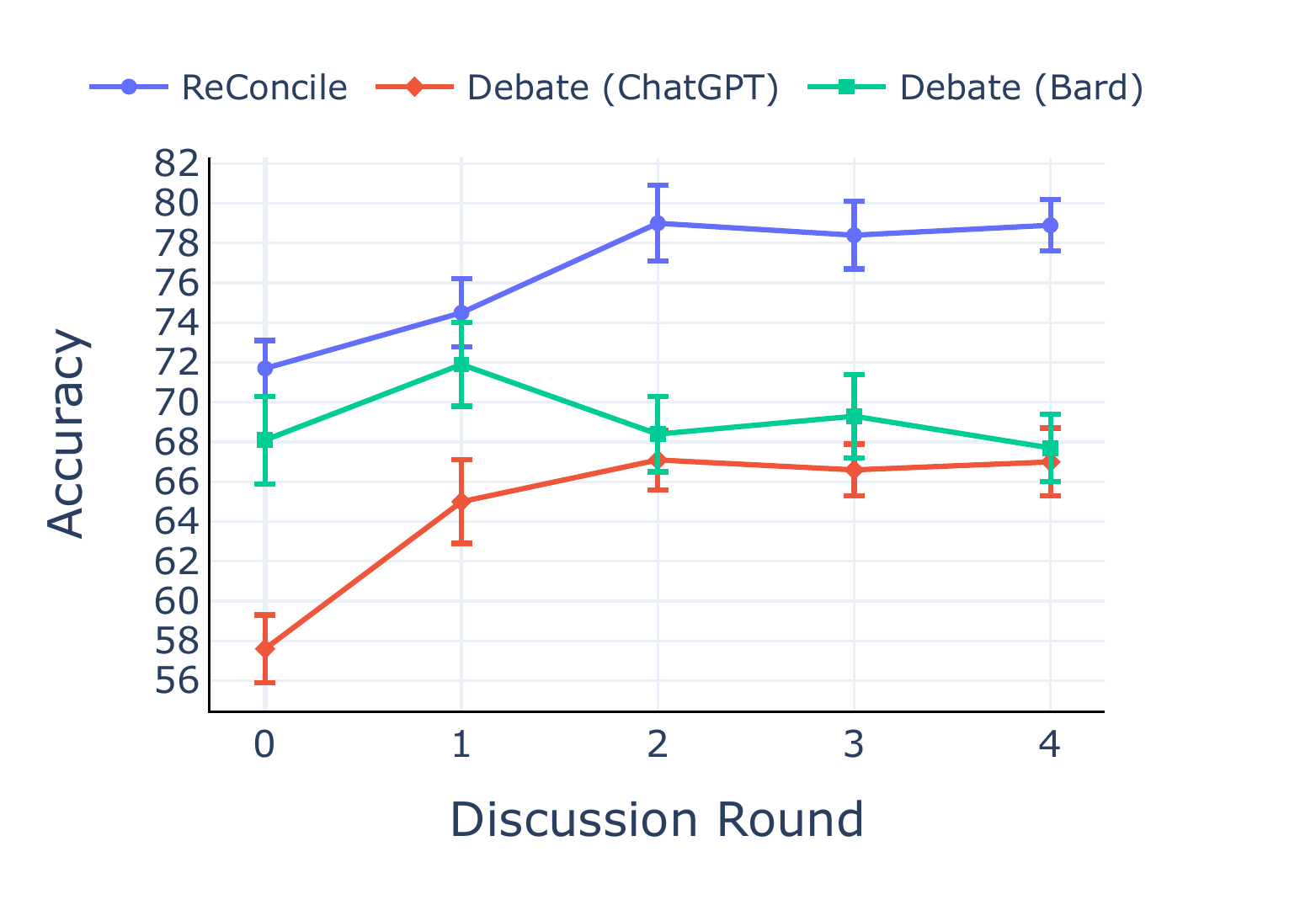}\label{fig:round_part1}} 
    \subfigure[]{\includegraphics[width=0.329\textwidth]{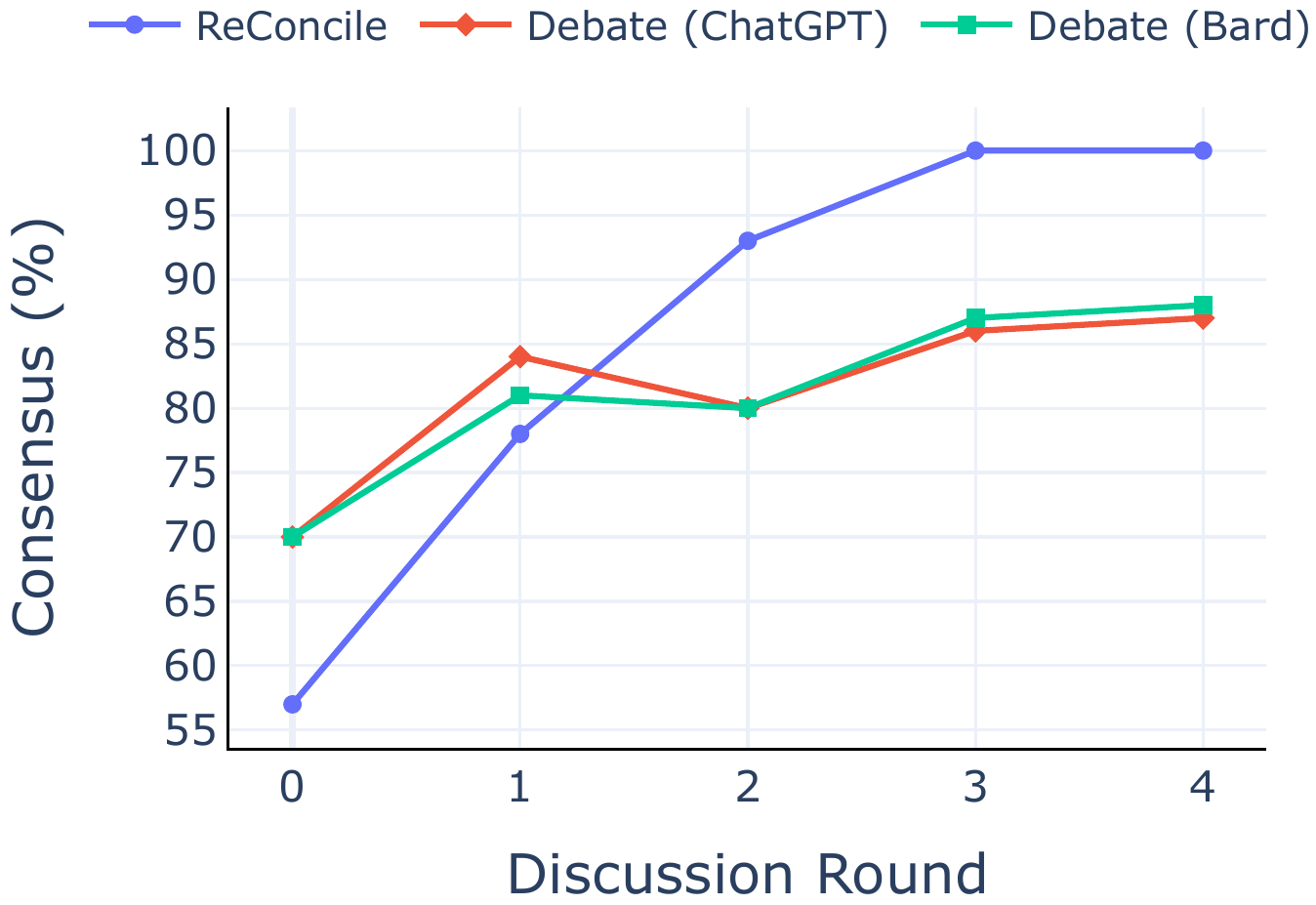}\label{fig:round_part2}} 
      \subfigure[]{\includegraphics[width=0.329\textwidth]{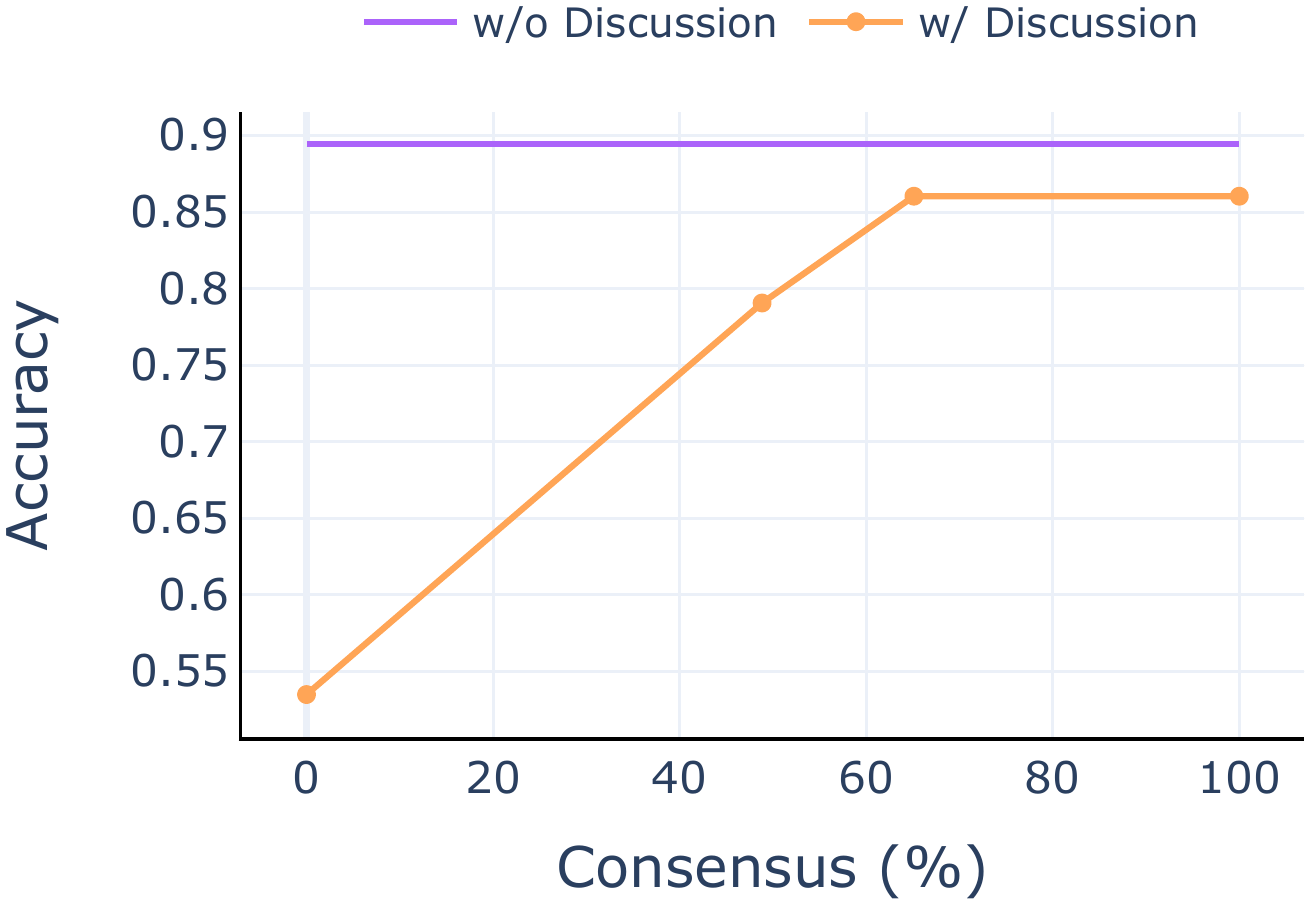}\label{fig:round_part3}} 
        \vspace{-10pt}

  \caption{\method{} achieves better and faster consensus. (a) Comparison of \method{} with Debate baselines showing the accuracy after each round. (b) Fraction of samples for which a consensus is reached after each round. (c) Accuracy as a function of consensus.
  }
  \label{fig:round}
\end{figure*}

\paragraph{Different models enhance response diversity.}
As was shown in Table~\ref{table:ablation}, \method{} obtains the most improvements via its \emph{multi-model} component. This surpasses \method{} with multiple ChatGPT instances, even when the generations sampled from these instances are encouraged to exhibit high diversity with a sufficiently high temperature. To further validate the importance of having multiple models and the diversity brought about by them, we develop a diversity metric. We hypothesize that if explanations from different models are indeed more diverse than those generated from multiple instances of the same model (e.g., in Multi-agent Debate), then our diversity metric should capture that. With that goal, we define diversity between multiple agents as the summation of the pair-wise diversity between agents: $D(A_1, A_2, A_3) = D(A_1, A_2) + D(A_1, A_3) + D(A_2, A_3)$, where $A_1$, $A_2$, and $A_3$ are the three agents' initial responses (either belonging to the same underlying model or different models). We then measure pairwise diversity by computing the cosine similarity between the response embeddings with BERTScore~\citep{zhang2019bertscore}. Note that lower similarity scores will mean greater diversity. With the diversity metric defined, we compute this metric for three variants: (a) paraphrased responses of a single ChatGPT to serve as a baseline, (b) responses from \method{} using three instances of a single ChatGPT model, and (c) responses from \method{} with ChatGPT, Bard, and Claude2 as agents. In Table~\ref{table:diversity}, we show that responses from different models exhibit the highest diversity (yielding the lowest similarity score of 0.8739) and also the highest accuracy (79.0\%), followed by the single-model variant (with a similarity score of 0.9102) and the paraphrased variant (with a similarity score of 0.9398). Thus, the higher diversity of (multi-model) \method{} means that agents have access to alternate solutions and external feedback, leading to better discussion and reasoning accuracy.  We also present a case study in Appendix~\ref{appendix:debate} to illustrate that the debate baseline sometimes struggles with echo chambers, stemming from a lack of external feedback, supporting the need for external feedback for improving LLMs~\citep{huang2023large}.

\paragraph{\method{} improves all agents individually.} We showed that the team performance of the agents improves through discussion. Next, in Table~\ref{table:round_acc}, we also present the accuracy of each agent after every round, as well as the overall team accuracy for StrategyQA. Evidently, the individual performance of each agent also improves alongside the team's performance. 

\begin{table}[t]
\small
\centering
\begin{tabular}{lcccccc}
\toprule
       Round & ChatGPT  & Bard  & Claude2   & Team \\\midrule
0 & 71.0\std{2.1}  & 71.7\std{0.9}    & 73.7\std{1.7}       & 74.3\std{1.2} \\
1 & 71.3\std{0.9}  & 77.7\std{1.2}    & 75.3\std{0.8}       & 77.0\std{0.9} \\
2 & 76.7\std{0.8}  & \bf 77.3\std{1.4}    & \bf 77.7\std{0.9}       & \bf 79.0\std{0.5}  \\
3 & \bf 77.0\std{0.9}  & 76.7\std{0.8}    & 77.0\std{1.2}       & 78.7\std{1.2} \\\bottomrule
\end{tabular}
\caption{\label{table:round_acc}The round-wise accuracy of ChatGPT, Bard, and Claude2 and their team performance (using weighted vote) on StrategyQA.}
\end{table}

\paragraph{\method{} Reaches Faster and Better Consensus.} \method{} terminates the discussion when a consensus is reached. More discussion rounds are costlier due to the increased API calls. Hence, achieving faster consensus while maintaining comparable accuracy gains is more efficient. To study this, in Fig.~\ref{fig:round_part1}, we plot the accuracy trends after each round; in Fig.~\ref{fig:round_part2}, we plot the fraction of samples for which consensus has been reached; and in Fig.~\ref{fig:round_part3}, we analyze accuracy as a function of consensus. From the first plot, we make two important observations: (1) \method{} improves accuracy for two rounds, following which the accuracy saturates, (2) Compared to the debate baselines, \method{} is not only superior after every round but also peaks at a highest accuracy of 79.0\% (vs 71.3\% for the baselines). Next, from Fig.~\ref{fig:round_part2}, our observations are also two-fold: (1) In the initial rounds (0 and 1), \method{}'s consensus percentage is lower because the discussion takes place between diverse LLMs. Diverse agents lead to more differences in opinions initially. (2) However, as the discussion proceeds, \method{} establishes consensus for all samples by round 3, while in the baseline, 13\% of the samples do not converge even after round 4. Finally, Fig.~\ref{fig:round_part3} shows that for the samples that enter the discussion phase (i.e., their initial answers did not have a consensus), accuracy is positively correlated with consensus. In other words, as a greater number of samples reach a consensus, accuracy proportionally improves. In summary, \method{} reaches \emph{faster} and \emph{better} consensus compared to baselines.
\section{Conclusion}
We presented \method{}, a multi-agent framework for reasoning with diverse LLM agents, engaged in multiple rounds of discussion via confidence estimation and generating explanations that can correctively convince other agents. \method{} demonstrated strong results on multiple reasoning benchmarks, consistently outperforming prior single-agent and multi-agent baselines and even improving upon GPT-4 on some benchmarks. 

\section*{Limitations}
For the API-based models used in \method{}, we note that we lack complete knowledge of the data that these models have been exposed to, and their scales in terms of parameters. Moreover, due to the API access, we do not possess complete control over their behavior. Depending on API-based models also necessitates the need to prompt these models to estimate their confidence. While this approach proves effective as evidenced by our results, we note that these estimates remain post-hoc in nature. Nevertheless, it is worth highlighting that these limitations could potentially be mitigated in the future should more open-sourced models emerge and demonstrate robust capabilities in adhering to long instructions. 

\section*{Acknowledgments}
We thank Peter Hase, Elias Stengel-Eskin, and the anonymous
reviewers for useful feedback and suggestions regarding
experiments for useful feedback and suggestions regarding experiments. This work was supported by NSF-CAREER Award 1846185, NSF-AI Engage Institute DRL-2112635, DARPA MCS Grant N66001-19-2-4031, Accelerate Foundation Models Research program, and a Google PhD Fellowship. The views contained in this article are those of the authors and not of the funding agency.

\bibliography{custom}
\bibliographystyle{acl_natbib}

\appendix
\clearpage
\begin{algorithm*}[ht]
\small
\caption{\method{}: A Group-Discuss-And-Convince Framework}\label{alg:RTC}
\begin{algorithmic}
\Require Test Problem $Q$, Discussion Rounds $R$, Agents $\mathcal{A} = \{A_i\}_{i=1}^{n}$, Convincing Samples $\mathcal{C} = \{C_i\}_{i=1}^{n}$
\Function{\method{}}{$Q, R, \mathcal{A}, \mathcal{C}$}
\State $r \gets 0$                         
\While{$r \leq R  \text{ and not } \textsc{Consensus}(Q, \{a_i^{(r-1)}\}_{i=1}^{n})$}
\State {$S \gets [], P \gets []$}
\ForEach {$A_i \in \mathcal{A}$}
\If {$r = 0$}
\State $P_I \gets (Q, \mathcal{C})$                               \Comment{Initial prompt consists of question and convincing samples}
\State $a_i^{(0)}, e_i^{(0)}, p_i^{(0)} \gets A_i(P_I)$             \Comment{Generate initial answer, explanation, and confidence}
\Else
\State $P_D \gets (Q, a_i^{(r-1)}, e_i^{(r-1)}, p_i^{(r-1)}, \mathcal{C})$     \Comment{Discussion prompt}
\State $a_i^{(r)}, e_i^{(r)}, p_i^{(r)} \gets A_i(P_D)$
\EndIf
\State $S \gets S + [a_i^{(r)}], P \gets P + [p_i^{(r)}]$         \Comment{Append each agent's answer and confidence}
\EndFor
\State {$\hat{a}^{(r)} \gets \textsc{WeightedVote}(S, P)$}      \Comment{Get team answer through a confidence weighted vote}
\EndWhile
\State \Return {$\hat{a}^{(r)}$}                                
\EndFunction
\end{algorithmic}
\end{algorithm*}

\section{Additional Details of \method{}}
\subsection{Implementation Details}
\label{appendix:implementation}
We provide more implementation details of \method{} in this section. During decoding, we set the temperature to 0.7 for ChatGPT and Bard and use the default setting for Claude2. All implementations involving ChatGPT are using \emph{gpt-3.5-turbo-0613} from Azure OpenAI.\footnote{https://oai.azure.com/} We retrieve results from Claude2 by posting requests to their webpage\footnote{https://claude.ai/chats}, and for Bard, we use \emph{chat-bison-001} from PaLM2 API\footnote{https://developers.generativeai.google/products/palm}. For each agent, we use four demonstrations of convincing samples.
In addition, we provide the workflow of \method{} in Algorithm~\ref{alg:RTC}. Required input contains a test problem $Q$, maximum number of discussion rounds $R$, $n$ agents $\mathcal{A} = \{A_i\}_{i=1}^{n}$, and convincing samples $\mathcal{C} = \{C_i\}_{i=1}^{n}$ for each agent. The output would be the team answer $\hat{a}^{(r)}$.
For the open-source models LLaMA2-70B and DeepSeekMath, we use four RTX A6000 GPUs, each with 48GB memory to generate output from them.
\begin{figure}[!h]
    \centering
    \includegraphics[width=\columnwidth]{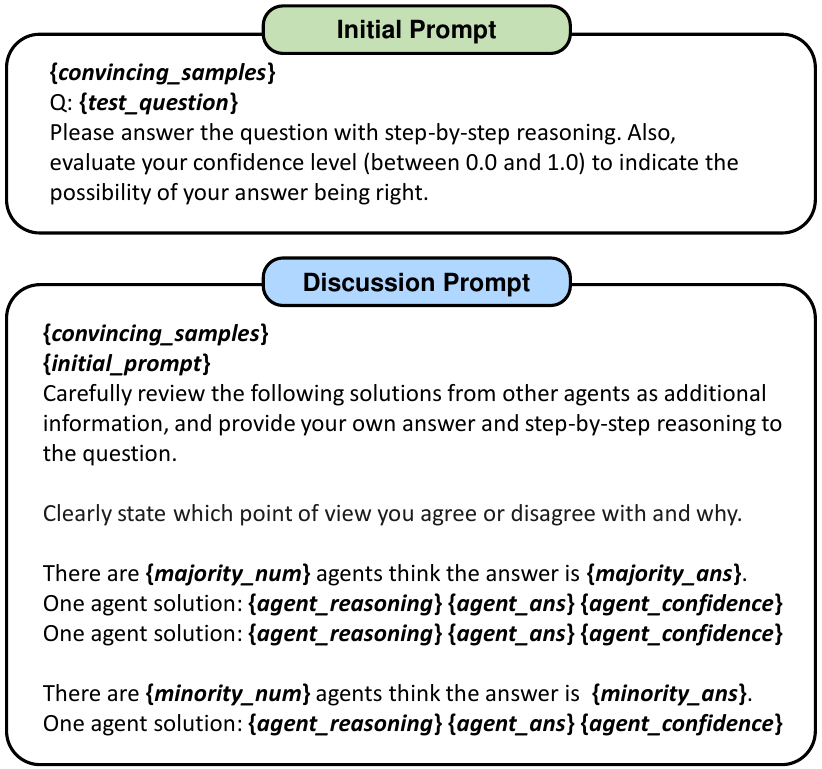}
    \caption{The prompts used in \method{} consist of an initial prompt and a discussion prompt.}
    \label{fig:prompt}
\end{figure}
\subsection{Initial Prompt and Discussion Prompt}
\label{appendix:prompt}
We show the prompts used in \method{} in Fig.~\ref{fig:prompt}. The initial prompt encompasses (1) the convincing samples that demonstrate how to convince other agents, (2) the test question, and (3) a requirement for `step-by-step' reasoning. The prompt also instructs the agent to express their confidence level, ranging from 0.0 to 1.0, indicating the likelihood of their answer being correct. The discussion prompt is an extension of the initial prompt, instructing the agent to review and express agreement or disagreement with other agents' solutions. To facilitate discussions, we design a grouping scheme that aggregates information based on the current opinions at the table. For instance, if two agents affirm that the answer to a given question is `yes' while the third agent disagrees with a `no', the designed grouping mechanism in the discussion prompt consolidates this information rather than simply concatenating all responses.

\section{Additional Results}
\label{appendix:results}

\subsection{Results on Full Test Sets}
\label{sec:full_test}
In Table~\ref{table:main_table}, we reported results with 100 test samples following several previous works and due to budget constraints. Upon experimenting on the full test sets of StrategyQA and Date Understanding, we confirm similar trends. Specifically, in Table~\ref{tab:full_test_set}, we compare \method{} to all of our major baselines and show that \method{} continues to outperform all baselines.
\begin{table}[]
\small
\centering
\begin{tabular}{lcc}
\toprule
\textbf{Model}     & \textbf{StrategyQA} & \textbf{Date} \\\midrule
ChatGPT            & 68.1                & 69.3                        \\
Bard               & 70.6                & 52.8                        \\
Claude2            & 72.7                & 77.9                        \\
Multi-agent Debate & 71.4                & 72.4                        \\
ReConcile          & \textbf{78.4}       & \textbf{84.5}               \\\bottomrule
\end{tabular}
\caption{Comparison of \method{} with baselines on the full test sets of StrategyQA and Date Understanding.}
\label{tab:full_test_set}
\end{table}

\subsection{Convincing Samples Improve Both \method{} and Multi-agent Debate}
\label{appendix:convincing_ablation}
\begin{table}
\small
    \centering
    \begin{tabular}{lcccccc}
        \toprule
        Method  & Accuracy \\ 
        \midrule
        Debate~\citep{du2023improving} & 66.7\std{3.1} \\
        RC (w/o Convincing Expl) & 74.5\std{1.7} \\
        RC (w/ Random Expl) & 75.0\std{2.5}  \\ 
        RC (w/ Convincing Expl) & 79.0\std{1.6}   \\ \midrule
        Debate (w/ Random Expl) & 68.7\std{2.2} \\
        Debate (w/ Convincing Expl) & 69.5\std{1.7} \\
        \bottomrule
    \end{tabular}
    \caption{Evaluation of the role of convincing samples on StrategyQA. \method{} (RC) without convincing samples outperforms multi-agent debate and with it obtains further gains. Convincing samples also boost the debate baseline.} 
    \label{table:convincingness_ablation}
\end{table}
Recall that \method{} selects a sample as convincing if the corresponding human explanation rectifies an agent's incorrect answer. Based on this, Table \ref{table:ablation} showed that by collecting only four human explanations, we can obtain significant improvements (`w/o Convincingness' row). Next, we consider a scenario where no human explanations are present. Table~\ref{table:convincingness_ablation} shows that even then, \method{} outperforms the debate baseline by absolute 7.8 points (second row). If random (i.e., general human explanations that may not necessarily ensure answer rectification)  are available (third row), we obtain some small improvements; but our convincing samples that are selected based on our novel answer-rectification criterion (fourth row) improve the results substantially. See  Sections~\ref{appendix:no_convincing} and~\ref{appendix:convincing_demo} for illustrative examples. Being able to convince another agent is also a generic concept that can be applied to other multi-agent systems, as demonstrated by improvements in the debate baseline (last row). 

\begin{table*}[]
\small
\centering
\begin{tabular}{lccccc}
\toprule
\textbf{Model}        & \textbf{StrategyQA} & \textbf{CSQA} & \textbf{GSM8K} & \textbf{AQuA} & \textbf{Date} \\\midrule
Claude2 (w/ 8-shot convincing samples) & 74.0\std{0.0}                & 69.7\std{1.2}          & 85.3\std{0.5}           & 64.3\std{1.2}          & 81.3\std{0.5}          \\
Self-Consistency w/ ChatGPT (9-way)           & 74.7\std{0.8}                & 73.3\std{1.2}          & \textbf{85.7}\std{0.4}            & 62.7\std{1.2}          & 70.3\std{0.9}          \\
\method{}      & \textbf{79.0}\std{1.6}       & \textbf{74.7}\std{0.4} & 85.3\std{2.2}  & \textbf{66.0}\std{0.8} & \textbf{86.7}\std{1.2} \\\bottomrule
\end{tabular}
\caption{Comparison of \method{} with Claude2 using 8-shot convincing samples and 9-way Self-Consistency.}
\label{tab:few_shot_comparison}
\end{table*}

\subsection{Comparison with Other Methods}
\label{sec:nine_way_SC}
In Table~\ref{tab:few_shot_comparison}, we compare \method{} to two other single-agent variants. While in our main Table~\ref{table:main_table}, we experimented with a random 8-shot Claude2 baseline, here we replace the in-context samples with our convincing samples. Even then, \method{} exhibits superior performance on all datasets except for GSM8K, again highlighting the importance of collaboration between diverse models. Next, we also report results for 9-way Self-Consistency which in terms of LLM calls represents the worst-case scenario of \method{} -- even for a more open-ended dataset like GSM8K, 9 LLM calls (i.e., 3 discussion rounds) happen in only 12\% of the samples and an even lesser 9\% on multiple-choice QA dataset like Date understanding. That said, \method{} continues to outperform 9-way SC by a large margin on most datasets.

\begin{table}[t]
\small
\centering
\begin{tabular}{lccc}
\toprule
&  Max Conf & Majority Vote &  Weighted Vote \\ \midrule
Accuracy  & 74.7\std{2.1} &  77.1\std{1.3}   & \bf 79.0\std{0.5}\\
\bottomrule
\end{tabular}
\caption{\label{tab:voting_strategies}Performance comparison of different voting strategies on StrategyQA. Weighted vote performs the best compared to simple majority vote and choosing the agent's answer with highest confidence.}
\end{table}

\begin{table}[t]
\small
\centering
\begin{tabular}{lcc}
\toprule
Voting weight & StrategyQA & GSM8K \\ \midrule
$w_1$            & 0.77       & 0.84  \\
$w_2$            & 0.79       & 0.83  \\
$w_3$            & 0.78       & 0.82  \\
$w_4$            & 0.77       & 0.83  \\
Majority      & 0.76       & 0.83  \\
Uncalibrated  & 0.78       & 0.84  \\
$w^*$ (Ours)          & \textbf{0.79}       & \textbf{0.85}  \\
\bottomrule
\end{tabular}
\caption{The robustness of the recalibation weight. We use the same weights $w^*$ across all datasets.}
\label{tab:voting_weight}
\end{table}

\begin{table}[t]
\small
\centering
\begin{tabular}{ll}
\toprule
\textbf{Dataset} & \textbf{License} \\ \midrule
StrategyQA & MIT License (\href{https://github.com/eladsegal/strategyqa/blob/main/LICENSE}{License}) \\ 
CommonsenseQA & MIT License (\href{https://github.com/jonathanherzig/commonsenseqa/issues/5}{License}) \\ 
GSM8K & MIT License (\href{https://github.com/openai/grade-school-math/blob/master/LICENSE}{License}) \\ 
AQuA & Apache 2.0 (\href{https://github.com/google-deepmind/AQuA/blob/master/LICENSE}{License}) \\
MATH & MIT License (\href{https://github.com/hendrycks/math/blob/main/LICENSE}{License}) \\ 
Date & Apache 2.0 (\href{https://github.com/google/BIG-bench/blob/main/LICENSE}{License}) \\
ANLI & CC BY-NC 4.0 (\href{https://github.com/facebookresearch/anli/blob/main/LICENSE}{License}) \\
\bottomrule
\end{tabular}
\caption{Dataset licenses}
\label{tab:dataset_licenses}
\end{table}

\subsection{Recalibration Strategy of \method{}}
\begin{figure*}[]
\centering
  \subfigure[]{\includegraphics[width=0.35\textwidth]{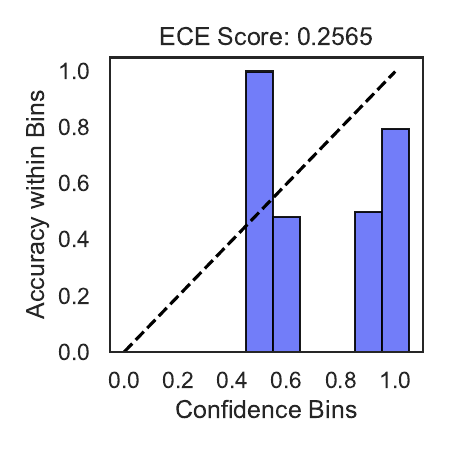}\label{fig:ece_part1}} 
    \subfigure[]{\includegraphics[width=0.35\textwidth]{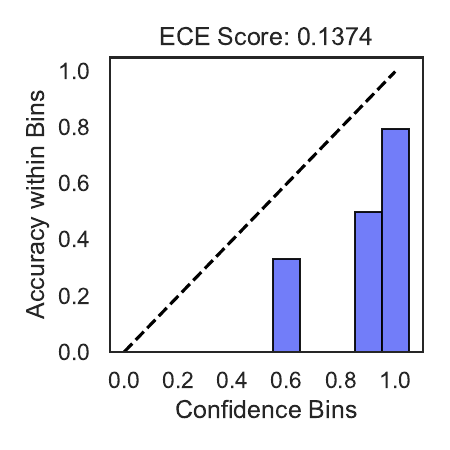}\label{fig:ece_part2}} 
  \caption{Expected Calibration Error (ECE)~\citep{naeini2015obtaining, guo2017calibration, 253510101} (a) before and (b) after confidence rescaling in \method{}. We observe a significant drop in ECE, showing the effectiveness of our simple method.}
  \label{fig:ece}
\end{figure*}

Directly using confidence scores as the voting weights is less effective due to the overconfidence problem of LLMs~\citep{xiong2023can,tian2023just,mielke-etal-2022-reducing}. Specifically, LLMs tend to produce consistently high confidence scores, which can make it challenging to discern subtle distinctions in confidence levels across different outputs. To address this, we employ a simple yet effective rescaling technique, facilitating better differentiation of confidence levels. This is expressed as:
\begin{equation}
\label{eq:weighted_vote}
    f(p_i^{(r)}) = 
\begin{cases}
    1.0,& \text{if } p_i^{(r)} = 1.0\\
    0.8,& \text{if } 0.9 \leq p_i^{(r)} < 1.0\\
    0.5,& \text{if } 0.8 \leq p_i^{(r)} < 0.9\\
    0.3,& \text{if } 0.6 < p_i^{(r)} < 0.8\\
    0.1,& \text{otherwise}
\end{cases} \nonumber
\end{equation}
where $p_i^{(r)}$ is the original confidence of agent $A_i$ in round $r$ and $f(p_i^{(r)})$ is the corresponding adjusted score.
To decide the optimal weights, we compare with a variety of settings including the majority vote and the uncalibrated confidence-weighted vote. The results are summarized in Table~\ref{tab:voting_weight}.
We denote the weight we used in our main experiment as $w^*=[1.0, 0.8, 0.5, 0.3, 0.1]$ where each value corresponds to the recalibrated confidence score. We further compare with other settings:
\begin{itemize}[noitemsep, topsep=0pt, parsep=0pt, partopsep=0pt]
    \item $w_1 =  [1.0, 0.9, 0.7, 0.5, 0.3]$
    \item $w_2 =  [1.0, 0.9, 0.5, 0.3, 0.1]$
    \item $w_3 =  [1.0, 0.8, 0.6, 0.4, 0.2]$
    \item $w_4 =  [1.0, 0.75, 0.5, 0.25, 0.0]$
\end{itemize}
and the results show that our $w^*$ works the best across datasets. In our main experiment, we fix the weight using $w^*$ and it is constantly outperforming majority vote across all seven datasets.
In addition, Fig.~\ref{fig:ece} shows that it helps reduce the Expected Calibration Error (ECE), a popular calibration metric~\citep{naeini2015obtaining}. While we note that recalibration can also be achieved through a learned model (e.g., Platt Scaling~\citep{platt1999probabilistic}), we refrain from using such models because \method{} is primarily designed as a few-shot method, and developing a recalibration model would necessitate access to a substantial number of annotated samples. Therefore, we use $f(p_i^{(r)})$ to perform a weighted vote to generate the team answer.

\subsection{Comparison of Different Voting Strategies}
\label{sec:vote_strategies}
At the end of any round $r$, every agent in \method{} generates its answer. Here we explore three voting strategies: (1) maximum confidence vote, where the agent's answer with the maximum confidence score would be the final team answer, (2) unweighted majority vote, where each vote carries equal weight, irrespective of the confidence score, and (3) weighted vote, where we use the recalibrated confidence scores as the voting weights. As shown in Table~\ref{tab:voting_strategies}, weighted vote is the most effective way to aggregate the team answer.

\section{Qualitative Examples}
\subsection{Convincing Samples for Each Agent}
\label{appendix:concincing_demo}
Table~\ref{tab:concincing_demo} shows examples of convincing samples on StrategyQA for each agent.

\subsection{Effect of Convincing Samples}
Here, we provide qualitative examples of how convincing samples change the way each agent responds to the question. We compare the initial responses (of each agent) with and without convincing samples in Table~\ref{tab:convincing_ablation}.
\begin{figure*}[!h]
    \centering
    \includegraphics[width=1.0\textwidth]{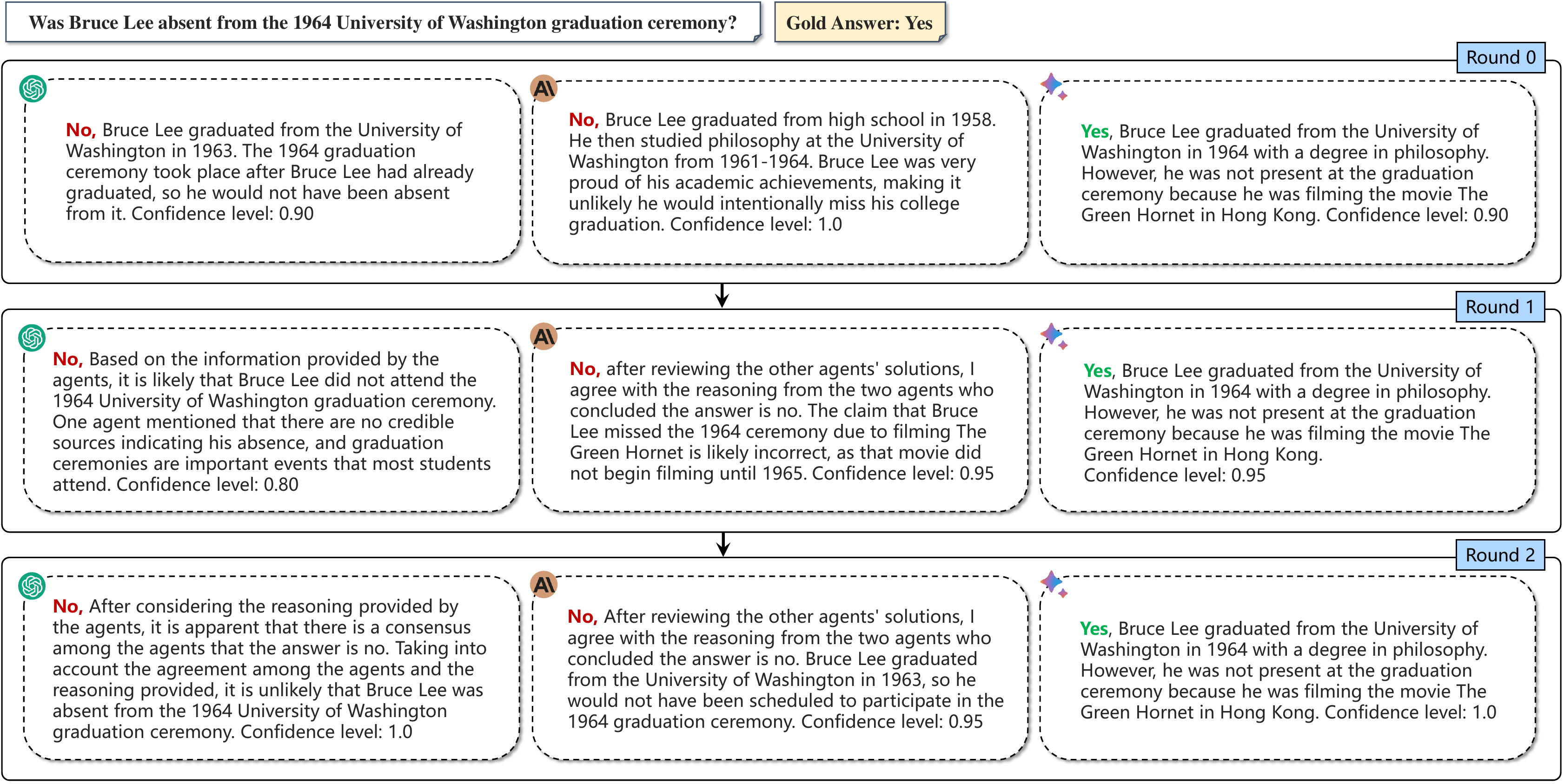}
    \caption{An illustrative example showing that without demonstrations of convincing samples, agents' opinions do not change much across the discussion rounds. All agents maintain their original stances and hence, never converge to the correct answer.}
    \label{fig:no_convincing}
\end{figure*}

\subsection{\method{} w/o Convincing Samples}
\label{appendix:no_convincing}
We notice that when \method{} operates in the absence of convincing samples, the agents tend to maintain their initial opinions more often. As depicted in Fig.~\ref{fig:no_convincing}, all three agents adhere to their original stances throughout the entire discussion and hence never converge to the correct answer.

\begin{figure*}[!h]
    \centering
    \includegraphics[width=1.0\textwidth]{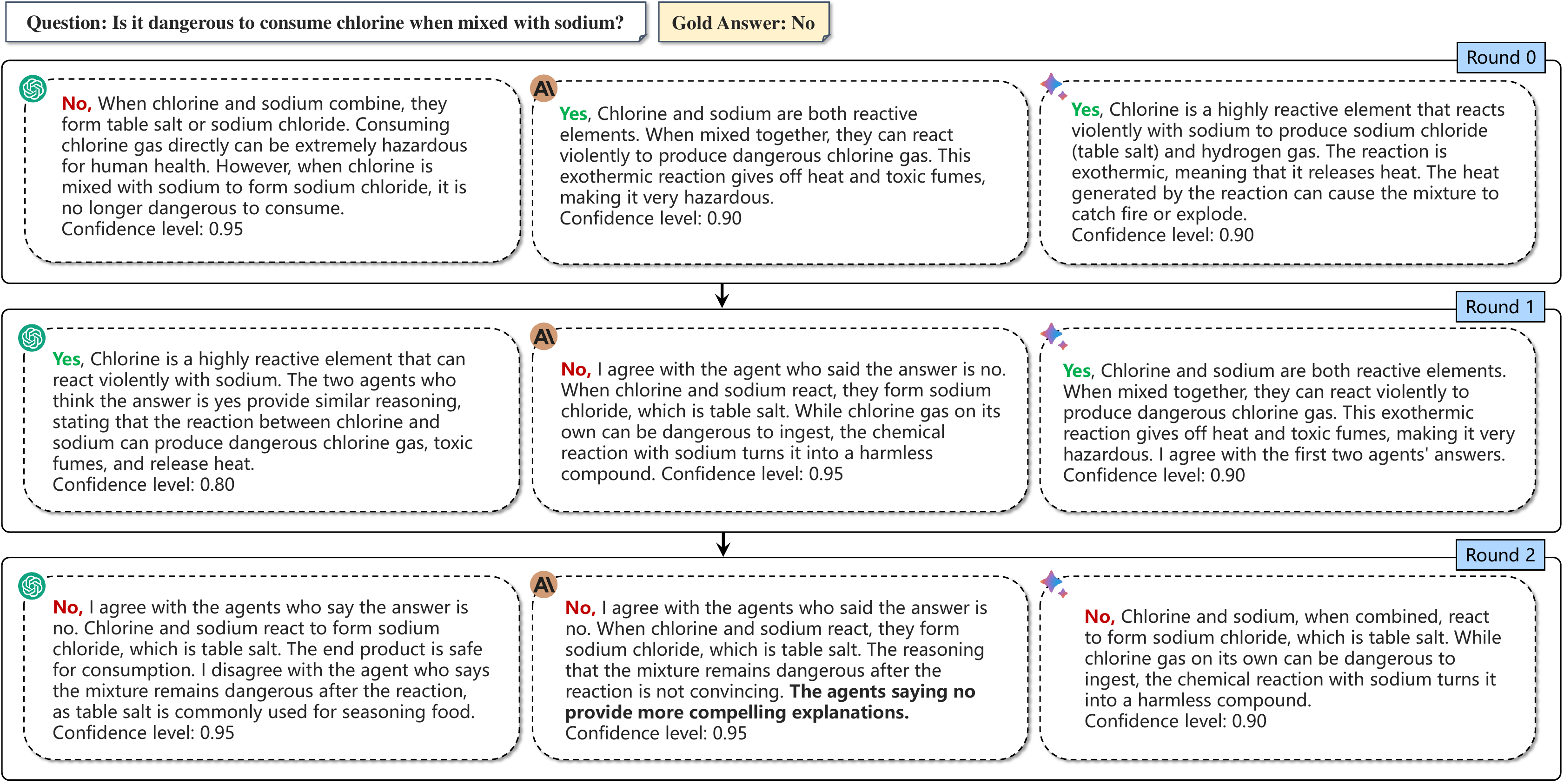}
    \caption{An illustrative example showing that with access to demonstrations of convincing samples, agents' opinions and explanations change during the course of multiple discussion rounds.}
    \label{fig:convincing}
\end{figure*}

\subsection{\method{} with Convincing Samples}
\label{appendix:convincing_demo}
On the contrary, when convincing samples are present, we show how the explanations of all agents change during the course of a discussion (see Fig.~\ref{fig:convincing}). Initially, Bard and Claude2 provide incorrect answers, but as the discussion unfolds, both agents revise their initial predictions, ultimately arriving at the correct answer.

\subsection{Single-Model Multi-Agent Debate Struggles with Echo Chamber}
\label{appendix:debate}
\begin{figure*}
    \centering
    \includegraphics[width=1.0\textwidth]{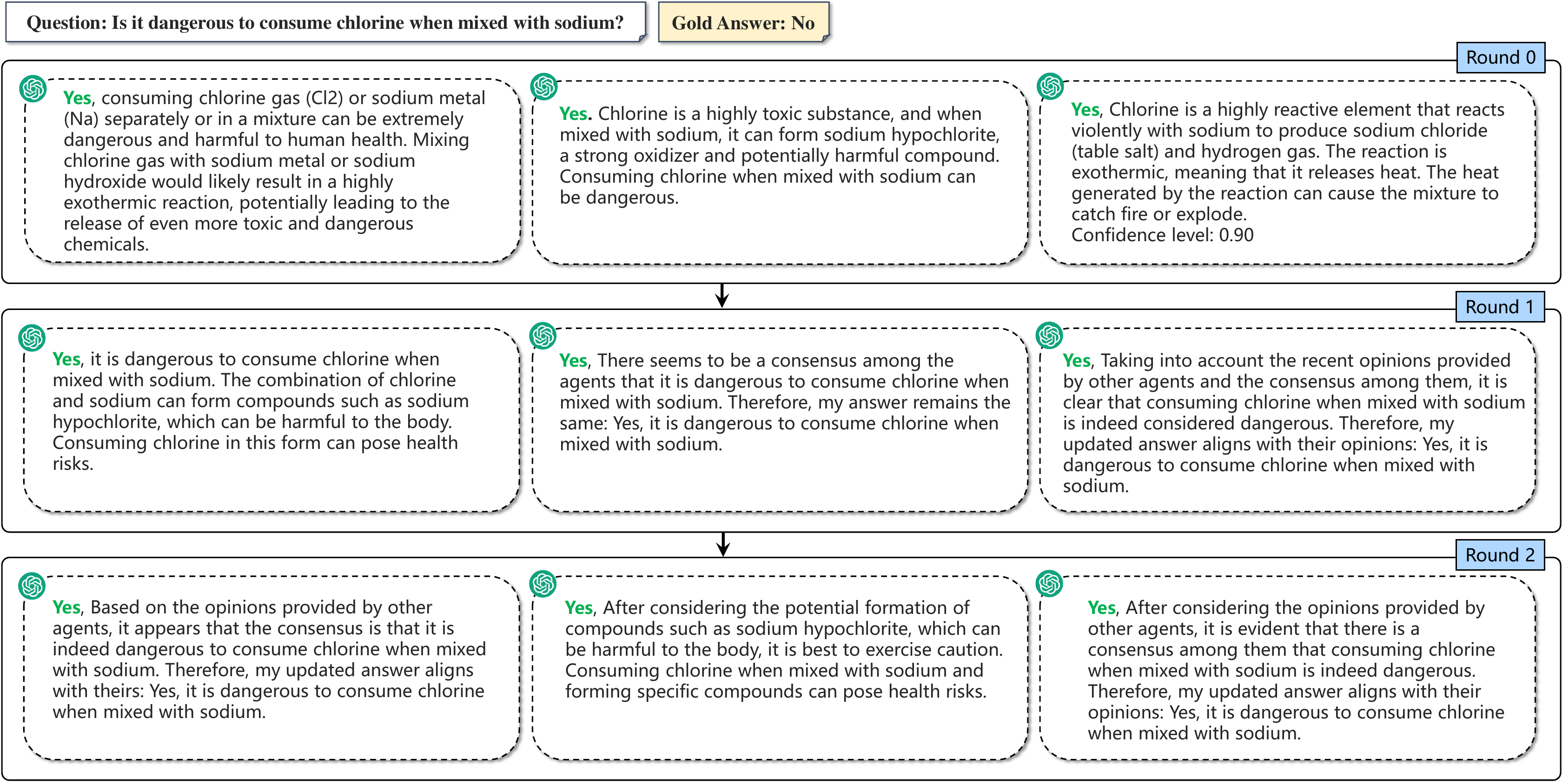}
    \caption{An illustrative example of multi-agent debate with multiple instances of ChatGPT. Initially, an incorrect answer is provided, and due to a lack of external feedback from diverse models, all agents persist with the same erroneous response throughout the debate process.}
    \label{fig:debate}
\end{figure*}

In Fig.~\ref{fig:debate}, we provide an illustration of multi-agent debate, implemented with multiple instances of the same underlying ChatGPT model. In this case, an incorrect answer is initially provided, but because external feedback from diverse models is lacking, all agents persist with the same incorrect response throughout the interaction.

\section{Benchmark Licenses}
The licenses of datasets we used are in Table~\ref{tab:dataset_licenses}.

\begin{table*}[!ht]
\begin{tabularx}{\textwidth}{X}
\Xhline{2\arrayrulewidth}
\\[-1.5ex]
\textcolor{MidnightBlue}{\textsc{Question}}: Is the tongue part of a creature's head? (Gold answer: \textcolor{ForestGreen}{\textbf{yes}})\\
\textcolor{WildStrawberry}{\textsc{ChatGPT (Initial)}}: The tongue is a muscular organ located in the mouth of most vertebrates, including humans. It is involved in functions such as taste, chewing, and speaking. The head, on the other hand, is the anatomical part that contains the brain, eyes, ears, nose, and mouth. Answer: \textcolor{BrickRed}{\textbf{no}}.

\textcolor{MidnightBlue}{\textsc{Convincing Human Explanation}}: A creature's tongue is inside its mouth. A creature's mouth is part of its head.

\textcolor{WildStrawberry}{\textsc{ChatGPT (w/ Convincing Human Explanation)}}:
Based on the information provided, the tongue is inside a creature's mouth. The mouth is part of a creature's head. Answer: \textcolor{ForestGreen}{\textbf{yes}}.
\\
\\[-1.5ex]
\hline
\\[-1.5ex]
\textcolor{MidnightBlue}{\textsc{Question}}: Are slime lilies in a different scientific family than asparagus? (Gold answer: \textcolor{ForestGreen}{\textbf{no}})\\
\textcolor{WildStrawberry}{\textsc{Bard (Initial)}}: Slime lilies are in the scientific family Araceae while asparagus is in the scientific family Asparagaceae. Answer: \textcolor{BrickRed}{\textbf{yes}}.

\textcolor{MidnightBlue}{\textsc{Convincing Human Explanation}}: Asparagus is a species of plants of the Asparagaceae family. Slime lilies are the common name for the flowering albuca plant. The albuca plant belongs to the scientific family of Asparagaceae.

\textcolor{WildStrawberry}{\textsc{Bard (w/ Convincing Human Explanation)}}:
Slime lilies are in the scientific family of Asparagaceae, while asparagus is in the same family. Answer: \textcolor{ForestGreen}{\textbf{no}}.
\\
\\[-1.5ex]
\hline
\\[-1.5ex]
\textcolor{MidnightBlue}{\textsc{Question}}: Would an astrologer focus on the densest terrestrial planet for a Friday horoscope? (Gold answer: \textcolor{ForestGreen}{\textbf{yes}})\\
\textcolor{WildStrawberry}{\textsc{Claude2 (Initial)}}: Astrologers do not focus on Earth for Friday horoscopes specifically. Astrologers tend to focus on the positions of the Sun, Moon, and planets in the zodiac for horoscopes. The day of the week is not especially relevant. Answer: \textcolor{BrickRed}{\textbf{no}}.

\textcolor{MidnightBlue}{\textsc{Convincing Human Explanation}}: Friday is associated with Venus in astrology Venus is the densest of the terrestrial planets

\textcolor{WildStrawberry}{\textsc{Claude2 (w/ Convincing Human Explanation)}}:
Friday is associated with Venus in astrology, and Venus is the densest of the terrestrial planets (Mercury, Venus, Earth, Mars). So the question is asking if an astrologer would focus on Venus for a Friday horoscope. Answer: \textcolor{ForestGreen}{\textbf{yes}}.
\\
\\[-1.5ex]
\Xhline{2\arrayrulewidth}
\end{tabularx}
\caption{Examples of convincing human explanations for each agent (ChatGPT, Bard, and Claude2). A convincing explanation is an instance of a answer-rectifying human explanation. For all three examples, the convincing explanation helps rectify an agent's initial incorrect answer.}
\label{tab:concincing_demo}
\end{table*}

\begin{table*}[!ht]
\begin{tabularx}{\textwidth}{X}
\Xhline{2\arrayrulewidth}
\\[-1.5ex]
\textcolor{MidnightBlue}{\textsc{Question}}: Was Lil Jon's top ranked Billboard song a collaboration with a member of The Lox? (Gold answer: \textcolor{ForestGreen}{\textbf{no}})\\
\textcolor{WildStrawberry}{\textsc{ChatGPT (w/o Convincing)}}: To determine if Lil Jon's top ranked Billboard song was a collaboration with a member of The Lox, we need to search for Lil Jon's discography and look for his highest-charting song on the Billboard charts. Then, we need to check if this song is a collaboration with a member of The Lox.
Answer: \textcolor{ForestGreen}{\textbf{no}}. Confidence: 0.8.\\
\textcolor{WildStrawberry}{\textsc{ChatGPT (w/ Convincing)}}:
Lil Jon's top ranked Billboard song is `Yeah!' which features vocals from Usher and Ludacris. There is no collaboration between Lil Jon and any member of The Lox on this song. Answer: \textcolor{ForestGreen}{\textbf{no}}. Confidence: 0.9.
\\
\\[-1.5ex]
\hline
\\[-1.5ex]
\textcolor{MidnightBlue}{\textsc{Question}}: Could ten gallons of seawater crush a six year old? (Gold answer: \textcolor{ForestGreen}{\textbf{yes}})\\
\textcolor{WildStrawberry}{\textsc{Bard (w/o Convincing)}}: Ten gallons of seawater weighs about 82 pounds. The average weight of a six year old is about 44 pounds. Therefore, ten gallons of seawater would be about twice as heavy as a six year old. \textbf{It is possible that ten gallons of seawater could crush a six year old, but it is also possible that the six year old could survive.} The outcome would depend on a number of factors, such as the size and strength of the six year old, the shape of the container holding the seawater, and the amount of time the six year old was submerged in the seawater. Answer: \textcolor{ForestGreen}{\textbf{yes}}. Confidence: 0.7.\\
\textcolor{WildStrawberry}{\textsc{Bard (w/ Convincing)}}:
The density of seawater is about 1.025 grams per cubic centimeter. Ten gallons of seawater is about 37.85 liters. The volume of a human body is about 60 liters. Therefore, ten gallons of seawater would be about 63\% of the volume of a human body. If ten gallons of seawater were to be poured on a six year old, it would be very difficult for the child to breathe and they would likely drown. Answer: \textcolor{ForestGreen}{\textbf{yes}}. Confidence: 0.95.
\\
\\[-1.5ex]
\hline
\\[-1.5ex]
\textcolor{MidnightBlue}{\textsc{Question}}: Would the top of Mount Fuji stick out of the Sea of Japan? (Gold answer: \textcolor{ForestGreen}{\textbf{yes}})\\
\textcolor{WildStrawberry}{\textsc{Claude2 (w/o Convincing)}}: To determine if the top of Mount Fuji would stick out of the Sea of Japan, I need to compare the elevation of Mount Fuji to the depth of the Sea of Japan. According to Wikipedia, the elevation of Mount Fuji is 3,776 meters. The average depth of the Sea of Japan is 1,752 meters according to Wikipedia. Since the elevation of Mount Fuji is greater than the average depth of the Sea of Japan, I can conclude that the top of Mount Fuji would stick out above the surface of the Sea of Japan if it was placed there. Answer: \textcolor{ForestGreen}{\textbf{yes}}. Confidence: 0.9.\\
\textcolor{WildStrawberry}{\textsc{Claude2 (w/ Convincing)}}:
Mount Fuji is the highest mountain in Japan, with a summit elevation of 12,389 feet or 3,776 meters above sea level. The average depth of the Sea of Japan is about 3,742 meters. Since the summit of Mount Fuji is higher than the average depth of the Sea of Japan, the top of Mount Fuji would stick out above the water if it was located in the middle of the Sea of Japan. Answer: \textcolor{ForestGreen}{\textbf{yes}}. Confidence: 0.9.
\\
\\[-1.5ex]
\Xhline{2\arrayrulewidth}
\end{tabularx}
\caption{Comparison of how each agent's reasoning changes with and without convincing samples. We observe that by conditioning on convincing samples, all agents tend to become more confident in their reasoning and generate less uncertain statements (shown in \textbf{bold}), which is also reflected in the actual confidence scores generated by each agent (e.g., goes up from 0.7 to 0.95 for Bard).}
\label{tab:convincing_ablation}
\end{table*}

\end{document}